\documentclass[sn-aps]{sn-jnl} 
 
\usepackage{graphicx}%
\usepackage{multirow}%
\usepackage{amsmath,amssymb,amsfonts}%
\usepackage{amsthm}%
\usepackage{mathrsfs}%
\usepackage[title]{appendix}%
\usepackage{xcolor}%
\usepackage{textcomp}%
\usepackage{manyfoot}%
\usepackage{booktabs}%
\usepackage{subfig}%
\usepackage{algorithm}%
\usepackage{algorithmicx}%
\usepackage{algpseudocode}%
\usepackage{listings}%
\usepackage{todonotes}
\usepackage{xcolor}
\usepackage{comment}

\theoremstyle{thmstyleone}%
\newtheorem{theorem}{Theorem}
\newtheorem{proposition}[theorem]{Proposition}%

\theoremstyle{thmstyletwo}%
\newtheorem{example}{Example}%
\newtheorem{remark}{Remark}%

\theoremstyle{thmstylethree}%
\newtheorem{definition}{Definition}%

\begin{document}

\title[Article Title]{Beyond Humans: Multispecies Animal Face Recognition Using Transfer Learning}


\author*[1]{\fnm{Maria} \sur{De Marsico}}\email{demarsico@di.uniroma1.it}

\author[2]{\fnm{Anil K.} \sur{Jain}}\email{jain@egr.msu.edu}
\equalcont{These authors contributed equally to this work.}

\author[3]{\fnm{Annalaura} \sur{Miglino}}\email{amiglino@unisa.it}
\equalcont{These authors contributed equally to this work.}

\affil*[1]{\orgdiv{Department of Computer Science}, \orgname{Sapienza University of Rome}, \orgaddress{\street{Via Salaria, 113}, \city{Rome}, \postcode{00185}, \state{State}, \country{Italy}}}

\affil[2]{\orgdiv{Department of Computer Science \& Engineering}, \orgname{Michigan State University}, \orgaddress{\street{Michigan State University Engineering Building 428 S. Shaw Lane}, \city{East Lansing}, \postcode{MI 48824 }, \state{Michigan}, \country{US}}}

\affil[3]{\orgdiv{Department of Computer Science}, \orgname{University of Salerno}, \orgaddress{\street{Via Giovanni Paolo II, 132}, \city{Fisciano (SA)}, \postcode{84084}, \state{State}, \country{Italy}}}


\abstract{The recognition of individual animals is relevant in several contexts. The search for lost or stolen pets, the re-identification and tracking of individuals of endangered species, and the recognition of animals in crowded breeding farms are among the possible applications.
Present recognition techniques mostly entail physical identification devices. The least invasive are normal collars, the most intrusive include means ranging from tattoos to ear tags to implanted microchips.
Each of them may have specific uses. Individuals of endangered species can hardly have a chip implanted, so physical tags or collars are used instead. However, they are often impractical and difficult to apply. Ear tags are often used in farms, while microchips are especially popular for pets. Both of them can be harmful. All of these means, more or less effective and more or less easy to counterfeit, could be substituted by remote recognition via the animal's face, which, if accurate enough, provides several advantages: it is non-invasive, can work at a distance, is difficult to counterfeit, as for instance in the case of the substitution of sick animals for healthy ones in the food industry. Of course, any recognition system needs appropriate training.
However, the few existing datasets suited to this goal, i.e., containing a sufficient number of per-subject images annotated with the single animal identity, are not large enough to train the current deep learning architectures. 
As an alternative, we investigate the possibility of exploiting pre-trained network models as backbones for transfer learning. 
Our experiments compared two models: the first one is FaceNet, specifically trained on large databases of human faces; the second is the Vision Transformer (ViT), pre-trained on ImageNet, i.e., on object categories. 
These approaches could be used for any species, given that they share a similar general facial structure. 
To demonstrate this, the experiments used three face datasets of very different animals: dogs, various endangered primates (lemurs, golden monkeys, and chimpanzees), and cattle. 
We report the results achieved and, for each dataset, compare them with the state of the art (SOTA), represented by ad hoc-trained deep networks, for the three recognition problems. 
It is interesting to note how the capture conditions differ among the three datasets, with a decrease in image quality (resolution, motion blur, diverse poses, etc.) from dogs to cattle to primates. This yields different results, even compared with the current SOTA.
The best performance was achieved with dog faces, where ViT reached a mean verification accuracy of 96.85\% and a Rank-1 Identification Rate of 84.34\%.
The results with endangered primates are still encouraging, but performance varies across animal classes and tasks (verification or identification), and, above all, does not always outperform SOTA. For cattle, the results outperform the state of the art when using ViT, while with FaceNet, they are still competitive.
}

\keywords{Transfer Learning, Animal Face Recognition, Dog Face Recognition, Primate Face Recognition, Cattle Face Recognition}



\maketitle

\section{Introduction}\label{s:intro}
Animals play different roles in our lives. Pets are part of our families and, as such, we love them and care for them. 
Unfortunately, adverse circumstances may arise that can break this relationship, such as a theft of valuable species or simply an escape. 
Data from the United States reports that approximately 80\% of lost domestic cats and dogs are never found, and lost pets account for a staggering one-third of the total\footnote{https://peeva.co/blog/missing-pet-epidemic-facts-and-figures}. To tackle this problem, a great contribution may come from the availability of a lost pet database for automatic search. 
The interest in the problem is testified by the dedicated sections of specialized sites like HumanePro, created in 2013 by John Polimeno, inspired by lost pet flyers on a Starbucks bulletin board\footnote{https://humanepro.org/ - Accessed February 2026}. The utility of related applications, also from a commercial perspective, is reflected in media articles.
For instance, the July 2018 ``Find Your Lost Pet With Facial Recognition"\footnote{https://idtechwire.com/find-your-lost-pet-facial-recognition-507106/ - Accessed February 2026} talks about the FindingRover app.  
The same app is also presented in the Animal Sheltering Magazine by HumanPro organization published in Spring 2020\footnote{https://humanepro.org/magazine/articles/face-value - Accessed February 2026}. The mobile app and website have been supported starting in 2017 by the Petco Foundation\footnote{The related site is https://lost.petcolove.org/}.
Another example still available online is the 2021 ``Petco Love Uses Pet Biometrics to Help Find
Missing Cats and Dogs" \footnote{https://idtechwire.com/petco-love-uses-pet-biometrics-help-find-missing-cats-dogs-042701/ - Accessed February 2026} 
announces that Iams, a well-known pet food company, launched NOSEiD, an app to identify lost dogs from their nose. 
%

%

The protection of endangered species through tracking individual animals to monitor their health and behavior covertly and possibly detect illegal transfers raises less sentimental but often more critical issues. Although less popular among consumer magazines and webzines, this problem calls for equally smart solutions. The International Union for Conservation of Nature (IUCN), which monitors the status of global biodiversity, publishes the annual Red List of Threatened Species and recently classified more than a quarter of mammal species as ‘critically endangered’, ‘endangered’, or ‘vulnerable’. Several species are in the limelight due to human activities, and some are considered symbols, like pandas, tigers, and polar bears, but about 60\% of species in the order Primates are threatened with extinction. Recognizing and tracking individual animals is a relevant activity for observing and understanding the processes affecting biodiversity and wildlife.

A further relevant field for animal recognition is the recognition and tracking of animals in large farms. In particular, the dairy industry struggles to enhance feed efficiency, detect and manage diseases in a timely and effective manner, and combat fraud involving meat from sick cattle placed on the market after falsified health documents.

At present, 'contact' means allow animal identification. The most well-known examples are microchips, ear tags, or tattoos. These can be painful and considered inhumane. A less hurting alternative is represented by collars, but they can be taken off or lost. It is also important to consider that, whatever the used identification, the capture of a wild animal, to allow starting tracking it by any device, can present difficulties, requires significant efforts, and causes potential harm. In this context, animal biometric recognition seems promising. 
The face is an especially suitable shared trait allowing distance and uncooperative capture.\\

The solutions to the extensively investigated problem of human face recognition 
achieve increasingly accurate results 
(see \cite{jain2011handbook} and the more recent \cite{wanyonyi2022open}). The recent advances of deep learning models have significantly improved performance (\cite{du2022elements}) with SOTA systems achieving above 99\% accuracy in controlled as well as semi-controlled conditions (see, for example, \cite{mishra2021multiscale} and \cite{Schroff_2015}).
%
%
%
%
%
The question raised by this work is whether advances in human face or object recognition could be applied to animal face recognition as well, without devising specific architectures. On the one hand, it is worth considering that
the well-known ``other race effect" \cite{otherraceeffect} studied for humans is even worse when the goal is to distinguish animals of the same breed or species. On the other hand, animal individuals belonging to different breeds, though in the same species, can differ in a way
that dramatically increases the variance to be reflected in a dataset. 
While the anatomical pattern of eyes-nose-mouth is shared by a lot of animals, it is sufficient to think of 
very different head shapes, ear styles, muzzle sizes, etc., which especially differentiate different breeds and present large variations. Figures \ref{fig:dogssamplevariance}, \ref{fig:lemurssamplevariance}, \ref{fig:goldensamplevariance}, \ref{fig:chimpsamplevariance}, and \ref{fig:cattlevariance} exemplify intra- and inter-individual variations in the animal datasets used here (see Section \ref{s:datasets}). In addition, possible differences in the quality of available images depending on the capture conditions raise further challenges. 
%
%
For instance, the best systems specifically trained for the recognition of dogs or endangered primate species stabilize at around 90\% accuracy \cite{app11052074}\cite{primates}.    
The main reason could be insufficient training data.
Most advanced techniques in face recognition are based on 
deep neural networks that need to be trained on a large number of faces of an even larger number of individuals. 
While there are plenty of human face datasets online, animal face datasets are very few and not equally large. 
%
The proposal evaluated here is to tackle the problem with a different strategy. The lack of data can be mitigated thanks to transfer learning. Using it, it is possible to take a network trained on a different task and adapt it to solve a (not too) different problem. With this approach, it is possible to use a smaller training dataset and take advantage of the knowledge developed on another one in a similar domain. 
Given that human face recognition is very accurate, especially in controlled conditions, the research hypothesis of this work is that the structural similarities (the triangle eyes-nose-mouth at least) of the faces of a high number of species (e.g., dogs, primates, and cattle considered here) can make the transfer learning approach very effective. In addition, recognition of thousands of object classes (see ImageNet \cite{imagenet_cvpr09}) has achieved very high performance.
Along these considerations, the presented experiments use and compare a very accurate face recognition network and a very accurate object recognition network, trained with such a huge amount of data that would be hardly available for the problem at hand. They are both fine-tuned and tested on the three datasets used, which present different image quality, to transfer the acquired knowledge to the presented case studies. This strategy should enable a more effective training of the final model with much less training data for animal recognition tasks. The experiments confirm that not only are performances generally better than the previous state of the art, at least for the cases of dogs and cattle, but further great advantages of the proposed method are its simplicity and cheapness in terms of data demand in the training phase. \\

\textcolor{black}{The paper extends the work in \cite{de2025adapting} by adding more discussion and presenting the results of the experiments on cattle face recognition.
The overall contributions are:}\\ 
1) to investigate the use of two different  Siamese networks to test transfer learning from either face recognition or object recognition models to dog face recognition and cattle face recognition; \\
2) to apply the same approach to a primate dataset with images of noticeably lower quality captured in more adverse conditions;\\
3) to compare the results with SOTA dedicated models.\\

\textcolor{black}{It is worth further underlining that the aim of the paper is not the proposal of a new architecture or the enhancement of existing ones.  It presents a preliminary feasibility study of using basic transfer learning from human face recognition models when not enough data is available to train deep learning architectures for animal face recognition. The description of the compared models in Section \ref{sec:compared} highlights how complex ad-hoc architectures may not achieve the same results of such a strategy.} 
%
It is also to note that the publicly available animal datasets can be divided into two categories. Some of them are rather devoted to species or breed recognition and therefore contain samples suitable for experiments on these different tasks. In other words, they hardly contain more than one image per individual, or at least, not a sufficient number to train over individual variations, and/or lack a suitable annotation for individual recognition. Others do not use animal faces but rather other anatomical characteristics, such as coat pattern (for cattle) or the muzzle/noseprint (cattle or dogs). \textcolor{black}{In addition, we selected datasets suitable to make a significant comparison with other works in the state of the art, i.e., those used in more works with the same goal.}\\

The paper continues as follows. 
Section \ref{s:related} introduces related literature. 
Section \ref{s:datasets} introduces the used datasets with their respective characteristics and the applied preprocessing. Section \ref{sec:compared} and \ref{sec:proposed} respectively introduce the compared architectures and the backbones used for the proposed transfer learning experiments. Section \ref{s:evaluation} details the experimental setup, designed to allow a fair comparison with reference works, and provides the detailed report of the recognition experiments divided by recognition type and dataset, with some final qualitative notes. Section \ref{s:conclusions} draws some conclusions and presents hypotheses for future work.
%
%
%
%
\section{Related Work}
\label{s:related}
The idea of using computer vision and machine learning for animal recognition is not new. 
The first applications of related methodologies aimed at supporting species identification 
for animal population studies in natural habitats. 
An example is \cite{wilber2013animal}, which proposes vision tools for field biologists that study animals in the Mojave Desert. The authors exploit a model for object recognition applied to the problem of species identification. This approach is similar to the one adopted here, yet at a different level of detail. They 
use transfer learning from object categories to species. We rather tackle the problem of recognizing single individuals, which requires, of course, finer discrimination abilities. This goal is more similar to that pursued by 
works dealing with animal re-identification, i.e., recognizing an individual upon re-encounter. These works are introduced in the following paragraphs. \\

The study of ecosystems' evolution, including the dynamics of a population of individuals, and behavioral ecology, relie on such ability \textbf{\cite{begon2020ecology}}. The most traditional methods entail tagging animals \textbf{\cite{lennox2023electronic}}, but camera traps or any image capture equipment in the environment are a viable and often better alternative \textbf{\cite{delisle2021next}}.  
Works using images, which can be captured without contact with the animal, exploit both machine learning algorithms fed with hand-crafted features and ad-hoc deep learning architectures. The experiments usually rely on datasets with a relatively reduced set of images and classes, both for training and testing. Furthermore, several strategies rely on comparing local characteristics like spots, scars, stains, cuts, or scratches, or the shape of a specific body part. 
The first use of computer vision techniques for animal re-ID can be found in a set of papers in the 1990 Report of the International Whaling Commission  
\cite{mizroch1990computer,hiby1990computer,whitehead1990computer} about recognition of whales. 
These mammals can be distinguished by characteristic features like a nick or a scratch. 
A similar strategy was used for elephants 
\cite{ardovini2008identifying}, using the shape comparison of the nicks characterizing the elephants’ ears. However, these features may not apply to all species. While they remain an important identification cue, the present work attempts to exploit results in human biometrics and general deep learning for object recognition to address the problem of the identification of animals presenting a face structure similar to humans. The interested readers can find an interesting and sufficiently complete survey of proposals for individual animal recognition in \cite{schneider2019past}.\\

    A problem that is common when starting to tackle the problem of individual animals recognition is that most publicly available animal datasets collect and annotate samples to train systems for species or breed recognition and not for individual animal recognition. Therefore, they contain a number of images per individual, when not a single one, that is not sufficient to train over individual variations. In addition, they are often not annotated with individual animal identity. Another common limitation is represented by the sample acquisition, which, if strictly controlled, is not suited to train systems to be possibly used in real-world applications. \textcolor{black}{A remarkable example of multi-family, multi-breed, multi-individual annotated dataset is PetFace \cite{shinoda2024petface}. It collects 257,484 unique individuals across 13 animal families and 319 breed categories, with a number of individuals per category ranging from 164,100 (cats) to 637 (chimps). All images are captured in quite controlled conditions.}

Regarding the animals considered in the presented experiments, a lemur dataset is presented in \cite{guan2023face}, but it was collected in controlled conditions. The lemur face identification for species protection is tackled in \cite{crouse2017lemurfaceid}.
%
The authors of \cite{primates} tackle the recognition of individuals in three different groups of primates, with a similar distribution of samples per individual as detailed in Section \ref{s:primates}. In the cited work, an ad-hoc deep architecture is trained and tested, demonstrating that it works better than transfer learning from a model trained on another animal.

The work proposing
DogFaceNet \cite{dogfacenet} for dog face recognition also presents the dog face dataset
that is used here. Before it, there was no open-source dataset with dog faces. 
%
The system proposed in \cite{dogfacenet} tackles the problem of dog face recognition with a new convolutional neural network architecture specifically created for this task. With this architecture, the authors claim to reach SOTA results on a version of the dataset smaller than the current one, with 92\%  in verification accuracy (verification on the current dataset is 86\% instead) and 60.44\% mean rank-1 accuracy for identification. However, dogs with fewer than 5 samples are discarded in order to achieve more generalizable results. In a following work \cite{app11052074}, 
all dogs are used instead, and this makes the original DogFaceNet performance decrease dramatically, with 76\% in verification accuracy (instead of 86\%) and a mean recognition rate of 10.96\% in identification (instead of 60.44\%). On the other hand, the cited research also explores a new training methodology, using a Vector Length loss besides the usual triplet loss, with outstanding results compared to the previous state of the art. Specifically, with this training method, the experiments achieve a verification accuracy of 88.8\% and a mean (over 100 iterations) recognition rate in identification of 34.92\% (impressive if compared with the previous SOTA value of 10.96\%). 
%

Cows2021 \cite{gao2021towards} is a large dataset of Holstein-Friesian cattle which is annotated by identity. It contains 10,402 labeled RGB images, captured to record the breed's distinctive black-and-white coat pattern, which is distinctive enough to recognize individual animals. Therefore, the animal faces are seldom visible, and the dataset is not suited for the aim of this paper. Some datasets focus on muzzle-based cattle recognition, but the entire cattle face is not always visible. Finally, some recent works have focused on individual recognition of cattle using facial features. In particular, an initial study from 2024 \cite{bakhshayeshi2023intelligence} introduced the Cattely-Cattle-Face-Images-Dataset of 2500 images from 50 dairy cows (50 images per individual), partly collected in a farm and partly from YouTube.
    This dataset is employed in two additional contributions, both published in 2025, in which different methodological approaches are proposed. In the first one \cite{hu2025angusrecnet}, the authors propose the AngusRecNet model, developed and evaluated on a proprietary dataset and subsequently tested on the Cattely dataset. In the second work, conducted largely by the same group \cite{hu2025two}, the authors present a Dual-channel fused MobileViT (DCFuseViT) network for individual recognition and further created a proprietary dataset. 
    The work presented in \cite {ahmed2024dataset} introduces and uses a dataset of 2,893 images taken from 459 distinct cattle, with a variable number of samples per individual (several animals have only from 1 to 3 images). 
    Later, a larger dataset was collected \cite{liu2025part} known as CattleFace2025 dataset. It includes 11,880 cattle face images of 574 Holstein Friesian and Simmental individuals. In this work, Cattely is used to be able to make more comparisons with papers in the state of the art. 

    The results of the cited projects on dog face recognition, together with those for the primate and the Cattely datasets, serve as a starting point for the comparisons in the present work.

\section{Datasets}
\label{s:datasets}
The choice of datasets used here is motivated by the goal of comparing the transfer learning approach across different animals and data of varying quality to better evaluate its effectiveness. As a matter of fact, the datasets differ significantly in absolute dimensions and in the distribution of samples per individual, as well as in image quality. \textcolor{black}{Images in DogFace \cite{dogfacenet} are of quite good quality, images in the Primate Dataset \cite{primates} are much more problematic, being often captured in the wild, and finally images in Cattely dataset \cite{bakhshayeshi2023intelligence} are extracted from videos, therefore presenting an intermediate quality.} \\
\textcolor{black}{The experiments did not take into account the huge PetFace dataset \cite{shinoda2024petface} for various reasons. The first one is that images are of generally good quality. The aim of the present study was to investigate the effectiveness of transfer learning with datasets containing images of varying quality, in order to explore the feasibility of the approach in uncontrolled or under-controlled applications, e.g., wild animals monitoring or automated cattle control. The second motivation was to enforce the real-world assumption that subjects involved in training could differ from those involved in testing, unless we assume the possibility of retraining or fine-tuning a system each time it is transferred to a different context. Since this is not the philosophy underlying the experiments in \cite{shinoda2024petface}, it would not have been possible to make a fair performance comparison. As a matter of fact, in the cited work, the only experiments involving subjects unknown during testing are those specifically aiming at discovering the unseen individuals. In the present investigation, all subjects involved in recognition during testing are unseen. Last but not least, one of the aims of the experiments was to compare the performance of complex, ad-hoc designed architectures for each case study, with a shared transfer learning strategy.}\\
\textcolor{black}{A further important point to take into account is the minimum number of samples per subject to consider for training. For the dogs dataset, individuals with fewer than 5 samples were not included in the experiments, both in the compared works and here.  As already underlined, we are not training a classifier but a feature extractor to recognize unseen individuals; in this condition, the more samples per individual in the training set, the better. This allows learning more stable and generalizable decision boundaries, i.e., a better way to pull samples for the same individual as close as possible in the embedding space, taking into account various covariates of intra-class variance,  while pushing samples from different individuals as far apart as possible, taking into account possible elements of inter-class confusion. The number of 5 samples is quite popular in the machine learning and deep learning communities; however, formal demonstrations regarding the minimum theoretical number of samples for traditional training can be found in \cite{golestaneh2025many}, while the fact that such a limitation also holds for few-shot learning is discussed in \cite{triantafillou2017few}.}
\subsection{Dog Dataset}
\label{s:dogs}
\begin{figure}[h!]
    \centering
    \includegraphics[scale=0.70]{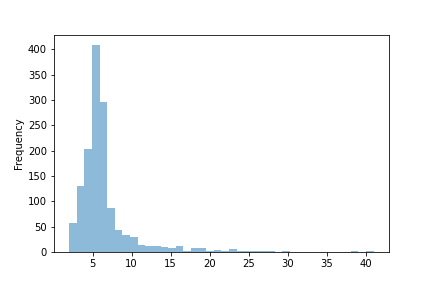}
    \caption{Dog dataset: histogram of no. of images per individual.}
    \label{fig:dogdist}
\end{figure}

\begin{figure*}[h!]
   \centering  
   \includegraphics[width=12cm]{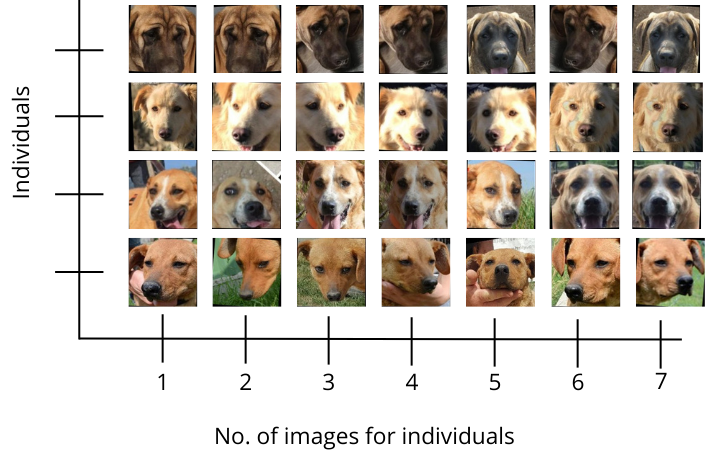}
    \caption{Examples of inter- and intra-individual sample variance in the dog dataset}
   \label{fig:dogssamplevariance}
\end{figure*}

The experiments in this work rely on a dataset of dog faces, which extends the original one collected and used with DogFaceNet \cite{dogfacenet}. The extension contains 8,363 images of 1,393 dogs (from 2 to 41 images per dog). The dataset is publicly available\footnote{Available at https://github.com/GuillaumeMougeot/DogFaceNet/releases/ - Accessed February 2026}.
Figure \ref{fig:dogdist} shows the frequency of the number of samples per individual. Figure \ref{fig:dogssamplevariance} shows examples of inter-individual (vertical) and intra-individual (horizontal) sample variations.

The dog face images are scraped from the internet. They are taken in loosely controlled conditions, entailing PIE (Pose, Illumination, and Expression) variations; however, their resolution and sharpness determine an overall good quality.
This \textit{in the wild dataset} is particularly 
suited for a real use application. 
Images collected in controlled conditions, meaning frontal pose and homogeneous illumination at least, would not really simulate a real use in this kind of recognition application, which involves animals photographed in spontaneous settings (and it would be really difficult to do otherwise). 
Each RGB image is normalized to $224 \times 224$ size.
The authors provide a train/test split. Using it and roughly 20\% of the training set for validation produces the following cardinalities:

\begin{enumerate}
   \item 6,128 training samples from 1,252 dogs
    \item 1,538 validation samples from 869 dogs
    \item 537 test samples from 93 dogs
\end{enumerate}
%
%

It is worth noting that the training and test sets have no overlapping identities (i.e., an identity-based partition). This increases the generalizability of the obtained models and
is particularly important to allow embedding them in a real-world consumer application. 
In that case, the models would probably be pre-trained on a training set including completely different animals from those that a user would add in a system gallery and/or use as probes in a search or to signal a lost dog (equivalent to the test set). In these cases, it may also happen that probes and the gallery are given by completely different users. Therefore, the system cannot and should not rely on the fact that identities in the training set appear in the test set.\\
\begin{figure}[h!]
  \centering
    \includegraphics[width=8.5cm]{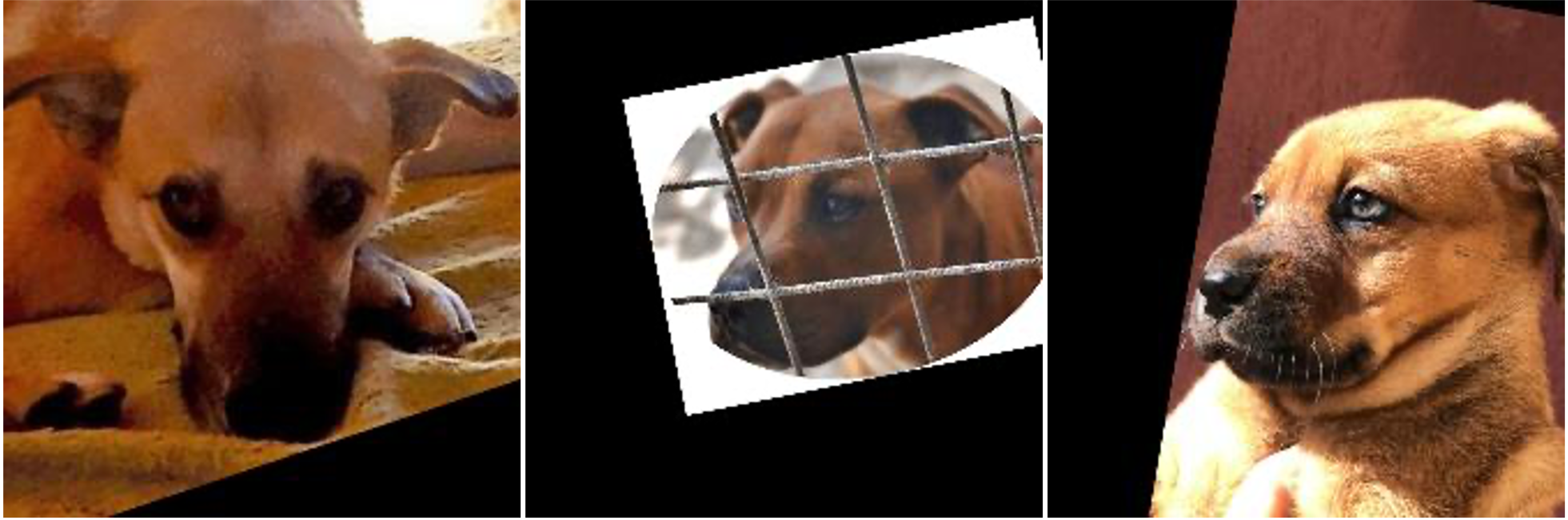}
    \caption{Example dog images in the dataset with black regions produced by the horizontal alignment of the eyes.}
   \label{fig:uncropped}
\end{figure}
%

%
%
\textbf{Original preprocessing}. 
 The images are already normalized for in-plane rotation. The black portion of the borders in Figure \ref{fig:uncropped} suggests the presence of counter-rotations to keep the eyes horizontally aligned in the upper part of the images and the muzzle a bit lower than the centers. 

\subsection{Endangered Primate Dataset}
\label{s:primates}
\begin{figure}[h!]
    \centering
    \subfloat[]{\includegraphics[scale=0.90]{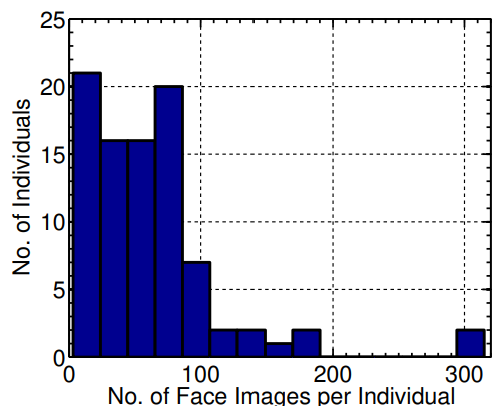}}\hspace{0.01cm}
    \subfloat[]{\includegraphics[scale=0.90]{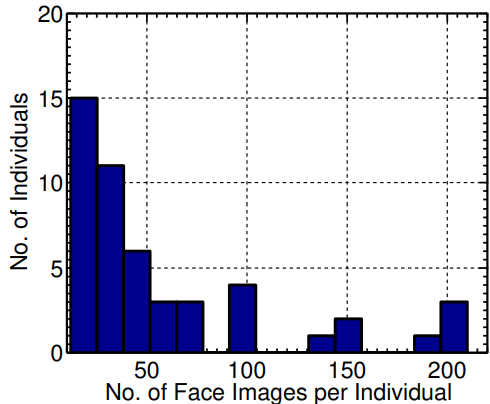}}\hspace{0.01cm}
    \subfloat[]{\includegraphics[scale=0.90]{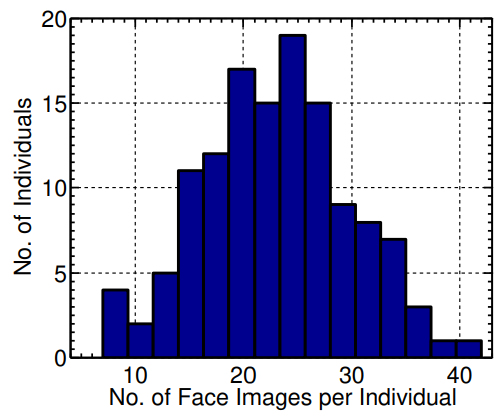}} \hspace{0.01cm}
    \caption{Primate dataset: histograms of no. of images per individual. Chimpanzees (a) Golden Monkeys (b) Lemurs classes (c).}
    \label{fig:prim_class}
\end{figure}

The Endangered Primate Dataset groups three collections with lemurs, chimpanzees, and golden monkeys. 
They are captured in similar conditions, which are much less favorable than those of the dog dataset.
The samples in each collection come from a single camera and are captured in the same place, either a national park or a zoo, depending on the collection. Images are smaller and more challenging than the dog dataset due to capture distance, resolution, and blur. Figure \ref{fig:prim_class} shows the histograms of the number of images per individual in the three collections\footnote{The dataset is available by contacting the authors.}.\\

\begin{figure}[h!]
   \centering  
\includegraphics[width=12cm]{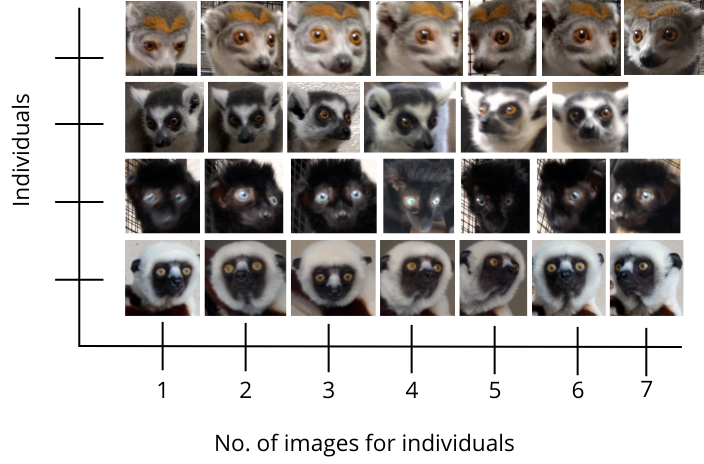}
    \caption{Examples of inter- and intra-individual sample variance in the lemurs.}
   \label{fig:lemurssamplevariance}
\end{figure}

\begin{figure}[h!]
   \centering  
\includegraphics[width=12cm]{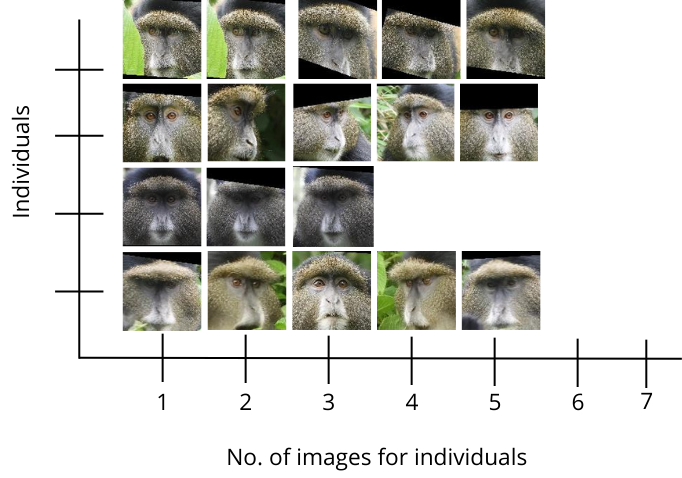}
   \caption{Examples of inter- and intra-individual sample variance in the golden monkeys.}
   \label{fig:goldensamplevariance}
\end{figure}

\begin{figure}[h!b]
   \centering  
\includegraphics[width=12cm]{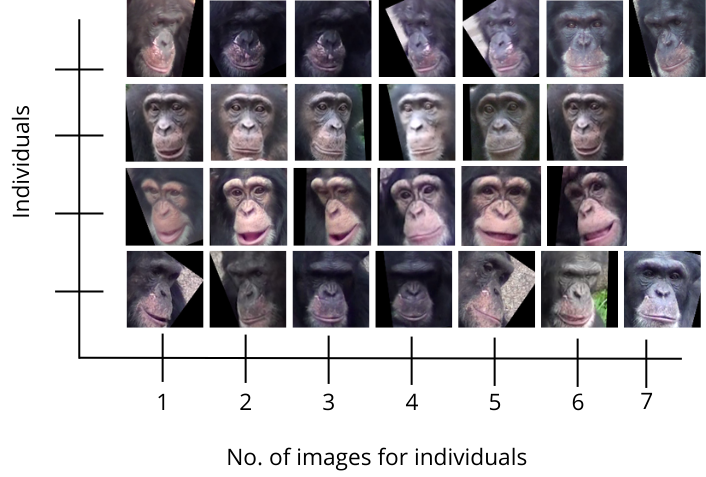}
    \caption{Examples of inter- and intra-individual sample variance in the chimpanzees.}
   \label{fig:chimpsamplevariance}
\end{figure}

%
\textbf{LemurFace Dataset}.
LemurFace contains 3,000 images of 129 lemurs. Images of lemurs were taken by one of the authors of \cite{primates} at the Duke Lemur Center (North Carolina, USA),  
using the 8 megapixel camera on a 2013 mid-range smartphone, an LG Nexus 5 
device\footnote{Details at https://www.gsmarena.com/lg\_nexus\_5-5705.php - Accessed May 2026}. \textcolor{black}{Therefore, they are quite low quality (4K photo/video resolution, 3264 × 2448 pixels)}. The fact that all the pictures were taken in the same environment could have raised some concerns about the capturing conditions and how they might influence the recognition process. To mitigate this issue, they were taken  
on two consecutive days, both indoors and outdoors. Fig. \ref{fig:prim_class}a shows the histogram of the number of samples per individual
and testifies to a high number of samples for most classes. 
Figure \ref{fig:lemurssamplevariance} shows examples of inter-individual (vertical) and intra-individual (horizontal) sample variations.\\
%
\textbf{GoldenMonkeyFace Dataset}.
GoldenMonkeyFace contains 1450 images of 49 golden monkeys.
One of the authors of \cite{primates} recorded 
241 videos in the Volcanoes National Park in Rwanda. 
using a Nikon Coolpix B700 \textcolor{black}{(20 megapixels, 5184 x 3888)}. 
Then the pictures were extrapolated from the video frames. However, the quality of the pictures is not that high, and this probably depends both on being frames from a video and on being shot from a distance with a telephoto lens. 
Fig. \ref{fig:prim_class}b shows the histogram of the number of samples per individual.
Figure \ref{fig:goldensamplevariance} shows examples of inter-individual (vertical) and intra-individual (horizontal) sample variations.\\
%
\textbf{ChimpanzeeFace Dataset}. 
ChimpanzeeFace combines two datasets, C-Tai and C-Zoo, presented by Loos and Ernst in \cite{LoosChimp} and then extended by Freytag et al. in \cite{freytag2016chimpanzee}. C-Zoo was collected in the Leipzig Zoo in Germany and contains 2109 faces of 24 individuals, \textcolor{black}{with a resolution of 6,403 megapixels}. C-Tai is captured in the Tai National Park in Cote d'Ivoire and contains 5078 face images belonging to 78 subjects, \textcolor{black}{with a resolution of 5,409 megapixels}. The final Chimpanzee dataset contains every subject except the ones with fewer than three samples, so the total is 5559 face images of 90 chimpanzees. 
Fig. \ref{fig:prim_class}c shows the histogram of the number of samples per individual.
Figure \ref{fig:chimpsamplevariance} shows examples of inter-individual (vertical) and intra-individual (horizontal) sample variations.\\
%
%
\textbf{Original Preprocessing}. 
\textcolor{black}{The three partitions of the primate datasets are already preprocessed. Aligning face images is a necessary preliminary step in most face recognition methodologies. This is done by detecting the face in the image and then the main face landmarks (eye corners, eye centers, mouth corners, etc.). Due to a lack of large primate face datasets for the three endangered species considered here, it is quite unfeasible to train a specialized face detector, especially given significant variations in fur and the low contrast between the eyes and the background. Therefore, all the face images in the three partitions of the dataset are manually annotated with three landmarks,
namely left eye, right eye, and mouth center. These landmarks
are used to construct a 'landmark template' used for face alignment before feeding the image to the system (more details in \cite{primates})}. 
Manually marked eyes and mouth landmarks allow aligning the images, which are
cropped around the face and rotated: 
the horizontally aligned eyes are on the top, and the mouth is on the bottom. 

%
%
%
%
%
%
%
%
%

\subsection{Cattle Dataset}
\label{s:cows}
\begin{figure}[h!]
    \centering
    \includegraphics[width=0.75\linewidth]{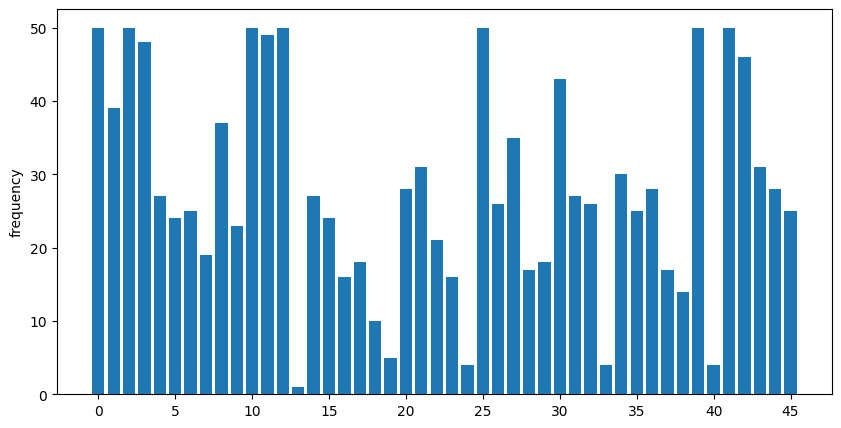}
    \caption{Cattle dataset: histogram of no. of images per individual.}
    \label{fig:cattledist}
\end{figure}

\begin{figure*}[h!]
   \centering  
   \includegraphics[width=12cm]{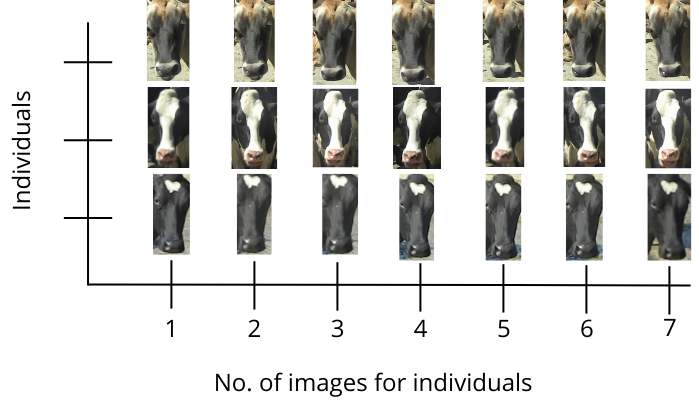}
    \caption{Examples of inter- and intra-individual face sample variance in the cattle.}
   \label{fig:cattlevariance}
\end{figure*}

The cattle Cattely dataset was introduced in \cite{bakhshayeshi2023intelligence}.\textcolor{black}{The full dataset contains frontal photos of 50 cattle with about 50 images each. Images were cropped from video recordings at Yattarna farm near Bodalla, New South Wales, and from related online YouTube videos. Only a subset of these images, known in literature as Cattely-Cattle-Face-Images-Dataset and used here, is publicly available\footnote{Available at https://github.com/aideep1400/Cattely-Cattle-Face-Images-Dataset - Accessed May 2026.}. It includes 392 images of 46 individuals. Images were resized to 1280x1280, and augmentation was applied to create 3 versions of each source image: random rotation between -5 and +5 degrees; random brightness adjustment between -25 and +25 percent. The total number of images, excluding the validation folder, is 1286. Even though larger cattle datasets are available, this is the one that appears more frequently in the literature and therefore allows a more extensive comparison.} Figure \ref{fig:cattledist} shows the histogram of the number of images per individual in this set of data, while Figure \ref{fig:cattlevariance} shows examples of intra-individual and inter-individual variations.
%
Images mainly depict two breeds: Holstein and Jersey.\\ 
%
\textbf{Original Preprocessing.} During preprocessing, face images were automatically aligned through the detection stage and subsequently resized and stretched to standard dimensions of 1280 × 1280 resolution. Additional preprocessing and augmentation steps, including rotation, brightness adjustment, rescaling, and color normalization, were applied to improve robustness to environmental variability and to reduce overfitting. Importantly, facial crops intentionally excluded the ears to prevent the model from exploiting ear tags as identity cues.

\section{The Compared Architectures}
\label{sec:compared}
\textcolor{black}{This section summarizes the main details of the compared architectures. We report them, at least in summary, to highlight the complexity of ad hoc recognition systems and to underline that, when the available training data is scarce, even the most sophisticated solutions may not achieve the best results. Interested readers can refer to the cited papers for further details.}

\begin{figure}[h!]
    \centering
    \includegraphics[width=1.00\linewidth]{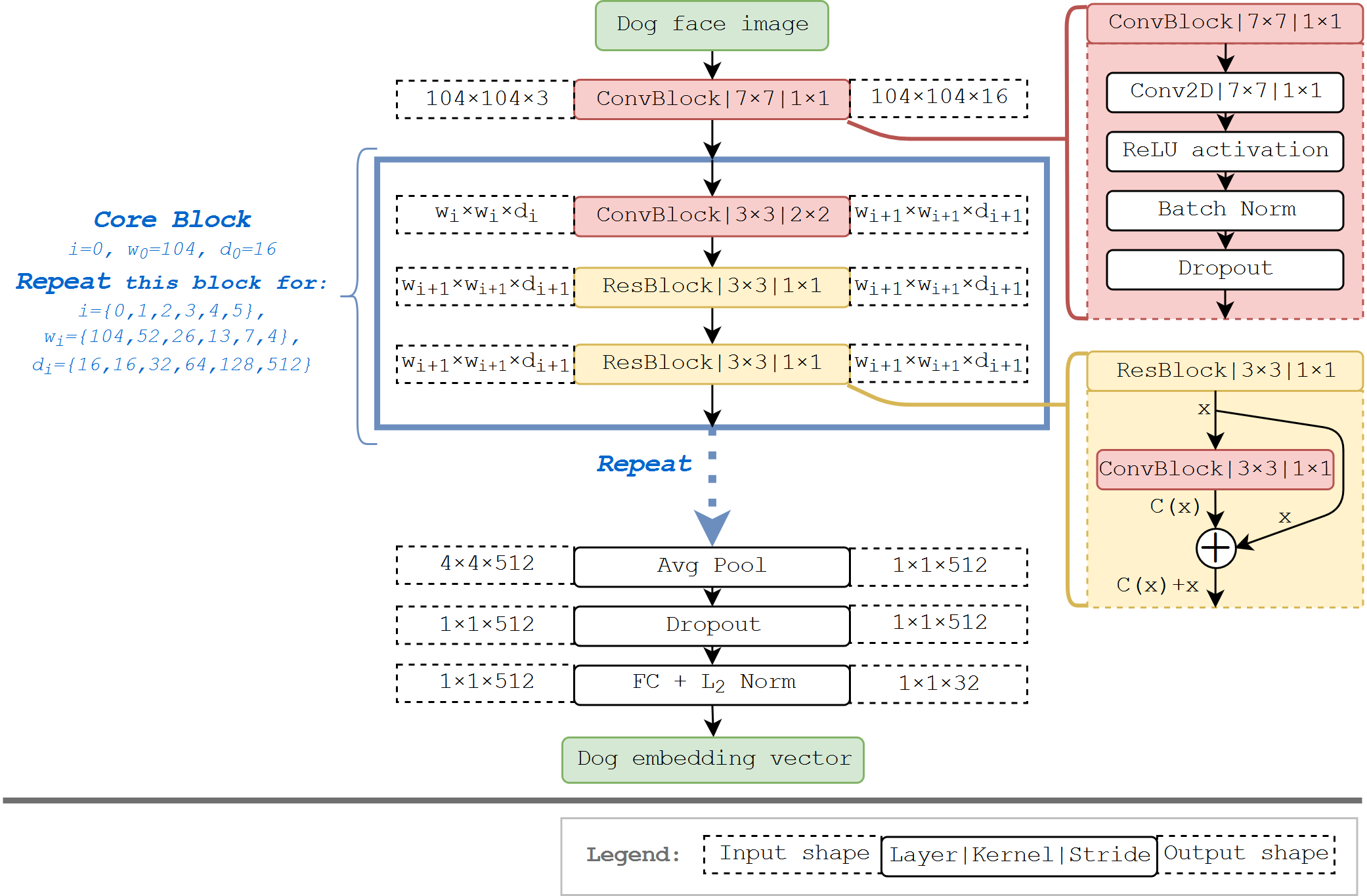}
    \caption{The architecture of DogFaceNet.}
    \label{fig:dogmodel}
\end{figure}

\textcolor{black}{\subsection{Dog Face Recognition}}
\textcolor{black}{The work presenting the DogFace dataset \cite{dogfacenet} proposes a dedicated architecture for dog recognition, namely DogFaceNet, which is based on ResNet. The network takes an image of size ($104 \times 104 \times 3$) as input and outputs
an embedding vector of size 32. 
The residual layers allow designing a very deep
network to extract as many features as possible. The final model (Figure \ref{fig:dogmodel})\footnote{The color image is published on https://github.com/GuillaumeMougeot/DogFaceNet/blob/master/images/model.png under MIT license.} contains 92 layers for a total of 5.8 million parameters. However, due to the small training set, the model would rapidly overfit; to avoid this, it also contains many dropout layers, each setting 75\% of the previous output to zero. Further limitation of overfitting is achieved using hard triplet mining, as in FaceNet \cite{Schroff_2015}. Hard triplets are generated offline, more specifically, every three
epochs, in order to make the computation lighter.}
\textcolor{black}{The architecture used in \cite{app11052074} is DogFaceNet, with the only difference being the two-stage training procedure that involves the specific Vector Length Loss. Usually, neural networks that output embeddings have a normalization layer at the end. This is the case of FaceNet: the L2 normalization layer helps the training by reducing
data sparsity by limiting the distribution of the embedding vector to the surface of a sphere with radius one. The idea behind the proposed loss function was rather to make the embedding space wider, allowing better data separation. The purpose of the Vector Length Loss (VLL) is to adjust the length of the embedding vectors in order to allow
the usage of the triplet loss without the normalization layer. In other words, the VLL aims to reduce the length difference between embedding vectors belonging to the same individual (or class) (see formal details in \cite{app11052074}). Unlike triplet loss, it does not use negative samples.  Regarding the two-stage training procedure, in the first stage, after producing the embedding vectors,   the model forks into two branches. One branch includes the L2-norm layer, and entails training in the usual way with L2 normalization followed by triplet loss. In parallel, the second branch entails training with VLL applied to the non-normalized embeddings. The two objective functions aim to minimize the two losses and train simultaneously the angle (triplet loss) and the length (VLL) of the embedding vectors. After stage 1, the embedding vector distribution lies within a certain area. In the second stage, the L2 normalization layer is removed, and only the triplet loss is used (with Euclidean distance instead of cosine), so that the embedding vector is
trained in a vector space rather than on the surface of a sphere. This method achieves better results than the pure DogFaceNet and will be used as the current SOTA reference.}

\begin{figure}[h!]
    \centering
    \includegraphics[width=1.00\linewidth]{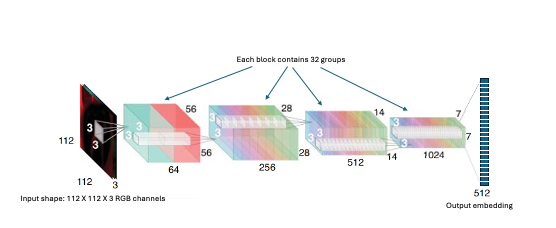}
    \caption{The architecture of PrimNet (adapted from \cite{primates}).}
    \label{fig:primnet}
\end{figure}

\textcolor{black}{\subsection{Primate Face Recognition}}
\textcolor{black}{The compared architecture for primate face recognition is PrimNet, introduced in \cite{primates}. This architecture modifies SphereNet-4 \cite{liu2017sphereface}, one of the smaller face recognition networks. Such architecture suffers severely from overfitting when trained on the relatively small primate datasets.
Hence, the authors of Primnet introduced two modifications to the SphereNet-4
architecture: they reduced the number of parameters by making the network
sparser through the group convolutional stratagem for all layers, as described in \cite{xie2017aggregated}, followed by channel shuffling \cite{zhang2018shufflenet}; at the same time, the number of channels is increased to make the network wider and improve the discrimination power of the hidden layers. In this way, Primnet becomes a sparse network, with a total of only 9.92 × 105 parameters, compared with Sphere-4, which has 1.26 × 107 parameters, and ShuffleNet, which has around 1.4 × 108 parameters. PrimNet is trained using
the AM-Softmax function, which has provided good performance in learning human face representations \cite{wang2018additive}. Figure \ref{fig:primnet} shows a sketch of the final architecture.}

\textcolor{black}{\subsection{Cattle Face Recognition}}
\textcolor{black}{One of the architectures that we compared with our approach for cattle recognition is taken from the paper introducing the cattle dataset \cite{bakhshayeshi2023intelligence}. The approach uses the Siamese Neural Network (SNN)
for one-shot learning (OSL). The paper compares the performance achieved by training an SNN with two different loss functions: the contrastive loss, which requires two branches, and the triplet loss, which requires three branches of CNNs to receive a set of three images (anchor, positive, and negative). In addition, different CNNs are used for the SNN branches. The best results (95.13\% accuracy on test data) are achieved with MobileNetV2 and triplet loss.}\\
\textcolor{black}{The other two architectures are introduced in \cite{hu2025angusrecnet} and \cite{hu2025two}. However, they are primarily designed for a more complex problem, i.e., the recognition of solid-colored cattle faces, where uniform color poses a source of confusion among individuals. In particular, the presented experiments use as main benchmarks two different datasets of Angus cattle, but also explore the performance of the proposed approaches on the same Cattely-Cattle-Face-Images-Dataset used for the present work.}\\
\textcolor{black}{The AngusRecNet architecture proposed in \cite{hu2025angusrecnet} processes Angus cattle facial images with a resolution of 640 × 640 over the three RGB channels. The Occlusion-Robust Feature Extraction Module (ORFEM) exploits
model-specific asymmetric convolutions to capture
global facial features, while small convolution kernels extract fine details.
The Vision AeroStack Module (VASM) further refines the extracted features by multi-scale spatial sampling.
In addition, AngusRecNet exploits the State-Space Dynamic Sampling Feature Pyramid Network
(SS-DSFPN) to model the dependencies between feature layers through dynamic sampling and feature fusion techniques. The detection head integrates the MCST-Head. The latter incorporates the CSFT (Channel Spatial Fusion Transformer) module to jointly model channel and spatial dimensions. It applies multi-head self-attention (MHSA) to the channel dimension to capture global dependencies between different channels and highlights key regions in the feature map through spatial convolution. The overall architecture is driven by high-level coordination of the different modules. During model training, AngusRecNet relies on online data augmentation in order to learn features under different interference/occlusion conditions. These techniques aim to simulate interferences on the Angus faces, such as feed and mud attachments. The augmentation is applied to the randomly selected 5 \% of the data in each training epoch. More details on the single modules can be found in \cite{hu2025angusrecnet}.}\\
\textcolor{black}{The same research group proposes in \cite{hu2025two} a two‑stage framework with feature‑contour and fine‑texture extraction, namely the EffiLibCatReID method. A Multi-Scale Edge Feature
Fusion Architecture (MSEFFA) is designed to strengthen sensitivity to edge information. The recognition stage of the proposed method adopts a dual-branch architecture, DCFuseViT. One branch extracts local contours and fine-grained textures, while the other captures
global structural information based on the MobileViT backbone.
The outputs of both branches are dynamically fused at
the feature level via a Local–Global Fusion Module (LGFM)
to obtain a unified and discriminative feature representation.
The fused features are further enhanced via a Squeezeand-Excitation (SE) residual connection. More details in \cite{hu2025two}.}\\

\section{The Proposed Transfer Learning Methodology}
\label{sec:proposed}
\textcolor{black}{The study aimed to investigate the feasibility of applying transfer learning from human face recognition to achieve good performance in animal face recognition. To this aim, the experiments compare transfer learning from two pre-trained models : (i) FaceNet 
\cite{Schroff_2015} for face recognition, which is based on InceptionResNetV1 \cite{szegedy2015going}, a hybrid network inspired by both Inception and ResNets, and represents a quite natural choice given the kind of items involved; (ii) a high-performance transformer model (see \cite{attention}) 
trained for general-purpose robust object recognition, namely the Vision Transformer (ViT) \cite{vit}. 
There are a few proposals for transformer-based face recognition (see \cite{nguyen2021clusformer} or \cite{jin2022avt}), but several pre-trained models for other tasks, such as general-purpose object recognition. The experiments investigated the possibility of transfer learning from one of them, as an alternative to a human face-specialized one. The experiments compare transfer from these two backbones and further compare the achieved performance with the dedicated architectures in the state of the art introduced in Section \ref{sec:compared}, explicitly designed and trained for the single case studies. It is worth anticipating that the adopted architectures are never used as classifiers, but rather as feature extractors, so that the returned embeddings are used to compare the input samples. Before introducing the choices adopted for the transfer methodology, it is worth reporting the additional preprocessing applied to the images besides the operations already performed on the image collections before distribution. \\}

\textcolor{black}{\subsection{Additional Preprocessing}}

\begin{figure*}[h!]
\centering
 \includegraphics[width=9.3cm]{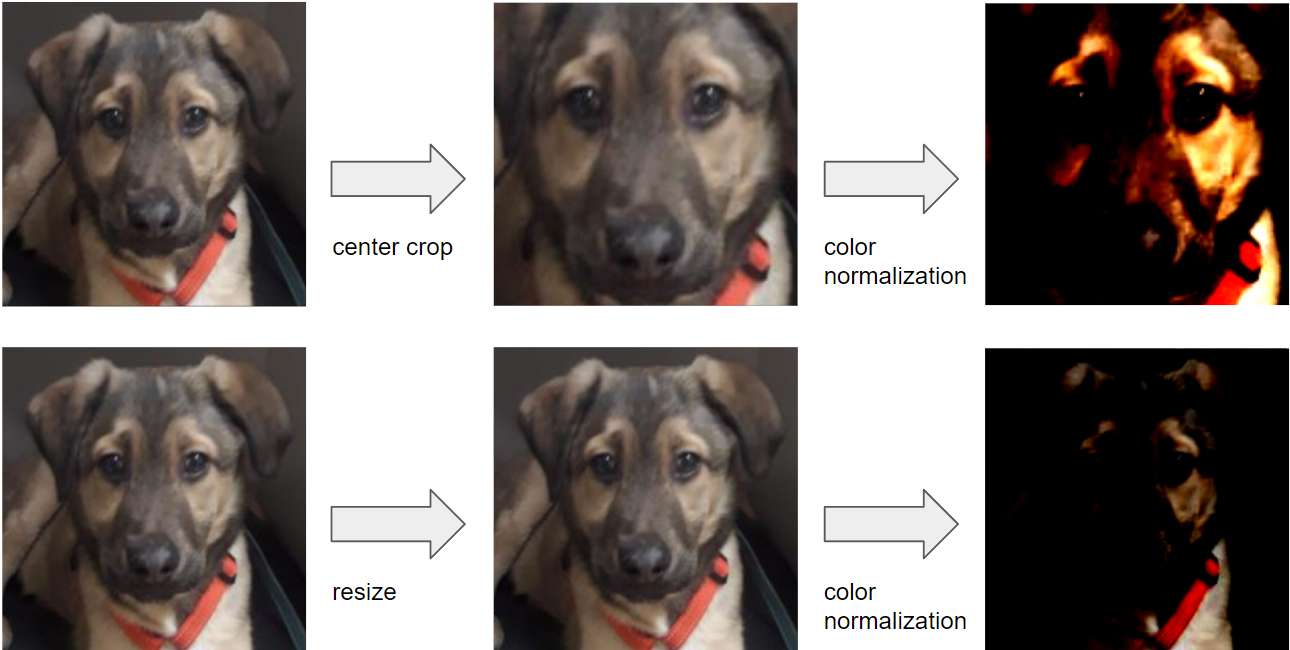}
   \caption{Example of the effects of an additional preprocessing pipeline for the dogs dataset for FaceNet (top) and ViT (bottom). }
    \label{fig:fullproc}
\end{figure*}

\textcolor{black}{Section \ref{s:datasets} presented the preprocessing operations already performed on the distributed image collections. 
A further model-dependent preprocessing step was applied to the datasets of dogs and primates for the reported experiments, but was not necessary for cattle, where it is part of the provided dataset preprocessing. Figure \ref{fig:fullproc} shows an example for the dogs dataset.\\}
%
\textbf{FaceNet}. 
In several images, the face is already centered, but the background takes a non-negligible amount of the whole picture. 
For FaceNet, the original preprocessing was extended by further cropping ($150 \times 150$) to reduce the amount of \textit{context} fed to the network. 
Of course, context could be useful, but deep networks tend to learn everything that recurs even if not relevant (e.g., in the case of dogs, the black portions of borders). Therefore, under some conditions, the model could focus more on the context than on the animal, or learn context elements as well. This may cause a false non-match just because the dog or other animal appears in a different scene. 
A further step of color normalization was separately applied by 
mean and standard deviation of the RGB channels.\\
%
%
%
\begin{equation}
   Norm\_C(x) = \frac{C(x) - \bar{(C_I)}}{\sigma(C_I)}
\end{equation}
%
where $x$ is a pixel in image $I$, $C$ stands in turn for $R$, $G$, and $B$ channels, while $\bar{(C_I)}$ and $\sigma(C_I)$ represent, respectively, the mean value and the standard deviation for channel $C$ over the whole image $I$.\\
%
%
\textbf{ViT}.
The ViT preprocessing follows the model pre-training \textit{vit-base-patch16-224}\footnote{https://huggingface.co/google/vit-base-patch16-224}. 
It includes image resizing ($224 \times 224$) and a color normalization with fixed values of 0.5 for both mean and standard deviation. \\

\begin{figure}[h!] 
  \centering
   \includegraphics[scale=0.29]{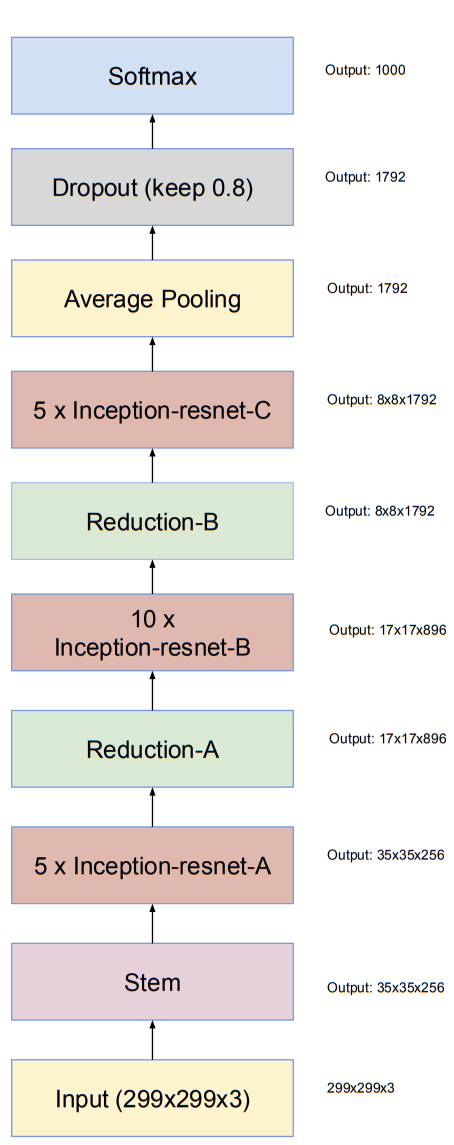}
   \caption{The full InceptionResNetV1 architecture}
   \label{fig:FaceNet}
\end{figure}

\textcolor{black}{\subsection{FaceNet}}
\textcolor{black}{The human face recognition model adopted for the presented experiments is FaceNet \cite{Schroff_2015}\footnote{The exploited implementation can be found on https://github.com/timesler/facenet-pytorch}. It is based on InceptionResNetV1 \cite{szegedy2015going}, a hybrid network inspired by both Inception and ResNets (Figure \ref{fig:FaceNet}). The authors claim 99.05\% accuracy when pre-trained on CASIA-Webface ($453,453$ face images of $10,575$ subjects) and 99.65\% accuracy when pre-trained on VGGFace2 ($~3.3M$ faces and $~9000$ classes). For the purposes of this study, preliminary tests demonstrated that the model trained on CASIA-Webface performs better. In more detail, InceptionResnetV1 is composed by a Stem phase, several InceptionResNet blocks with reductions in between, and finally a pooling, a dropout, and a softmax layer. 
}\\

\textcolor{black}{\subsection{ViT}}
\textcolor{black}{
Transformers are currently the most recent breakthrough in deep learning and they are virtually able to solve any task thanks to attention. In particular, the earliest architecture in this group was devised for language translation \cite{attention}. The Vision Transformer (ViT) was first introduced in \cite{vit} and is not very different from the original transformer architecture. The main difference between this and the original Transformer is that this architecture can take 2-dimensional images as input. The
original architecture of this standard transformer is suitable for fixed-size 1-dimensional sequences. Therefore, there is a preliminary
trainable linear projection layer that flattens the (non-overlapping) image patches, so transforming the 2D images into 1D vectors of fixed length, as needed by the transformer input. The outputs of this projection are called patch embeddings. }

\textcolor{black}{After the patch embeddings are created, a learnable embedding is prepended to the sequence, similarly to what happens with BERT \cite{devlin2019bert} class token. The state of this embedding at the transformer encoder's output serves as the image representation. In both phases of training (pre-training and fine-tuning) a classification head is attached to the output and it is implemented with a Multi-Layer Perceptron with a single hidden layer during pre-training and with just one linear layer during the fine-tuning phase. Position embeddings complement the patch embeddings in order to
retain positional information. Therefore, even though the image structure is lost
by flattening, the actual spatial information is still preserved. The position embeddings are just standard learnable unidimensional embeddings, as it seems that there is no
significant advantage in using more complex ones for 2-dimensional images. This
results in a sequence of embedding vectors that represent the input of the transformer
encoder. The transformer encoder is the usual alternating stack of multiheaded
self-attention layers and MLP layers, with layer normalization before every block.}

\begin{figure*}[h!]
    \centering
    \includegraphics[scale=0.5]{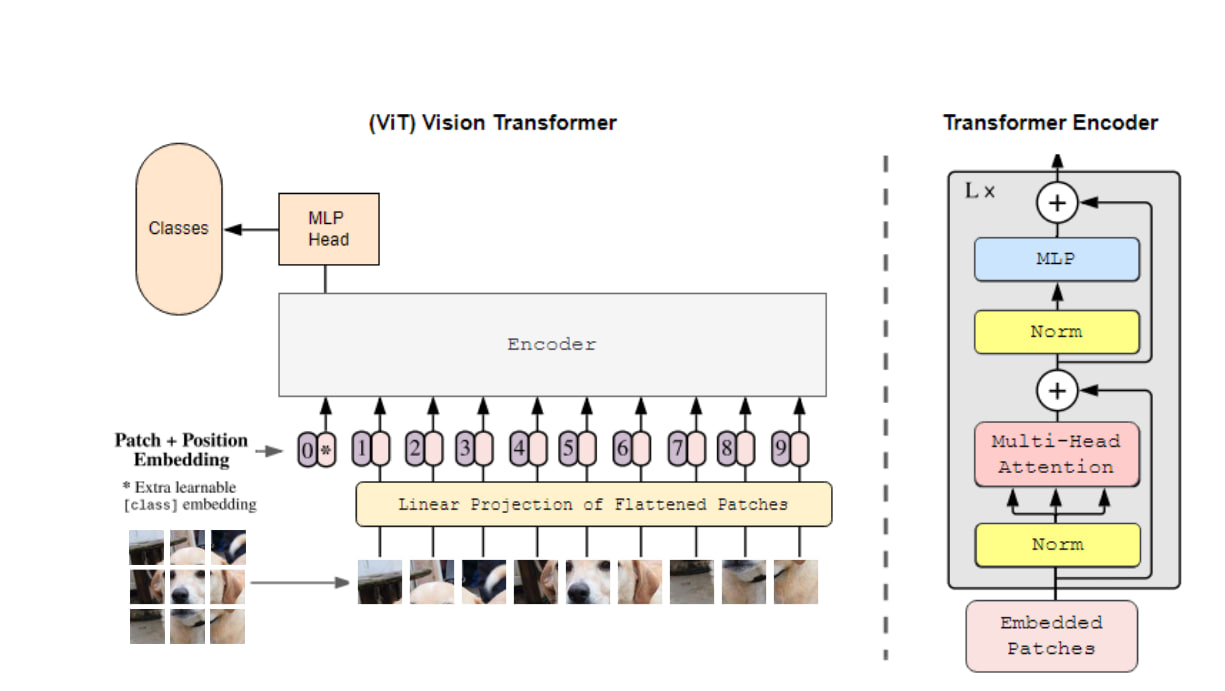}
    \caption{ViT architecture from the official paper \cite{vit}}
    \label{fig:ViTNet}
\end{figure*}

\textcolor{black}{
There are no pre-trained transformers for human face recognition, but many have been pre-trained on other tasks, such as general-purpose object recognition. That is the case of ViT. The model was pretrained on ImageNet-21k (21k classes and 14M images) in a supervised fashion, at a resolution of 224x224 pixels. Then the model was fine-tuned on ImageNet, a dataset containing 1 million images and 1,000 classes, also at a resolution of 224x224. 
Figure \ref{fig:ViTNet} shows the encoder made of alternating layers of multi-head self-attention. }\\

\begin{figure*}[h!]
   \centering
    \includegraphics[scale=0.5]{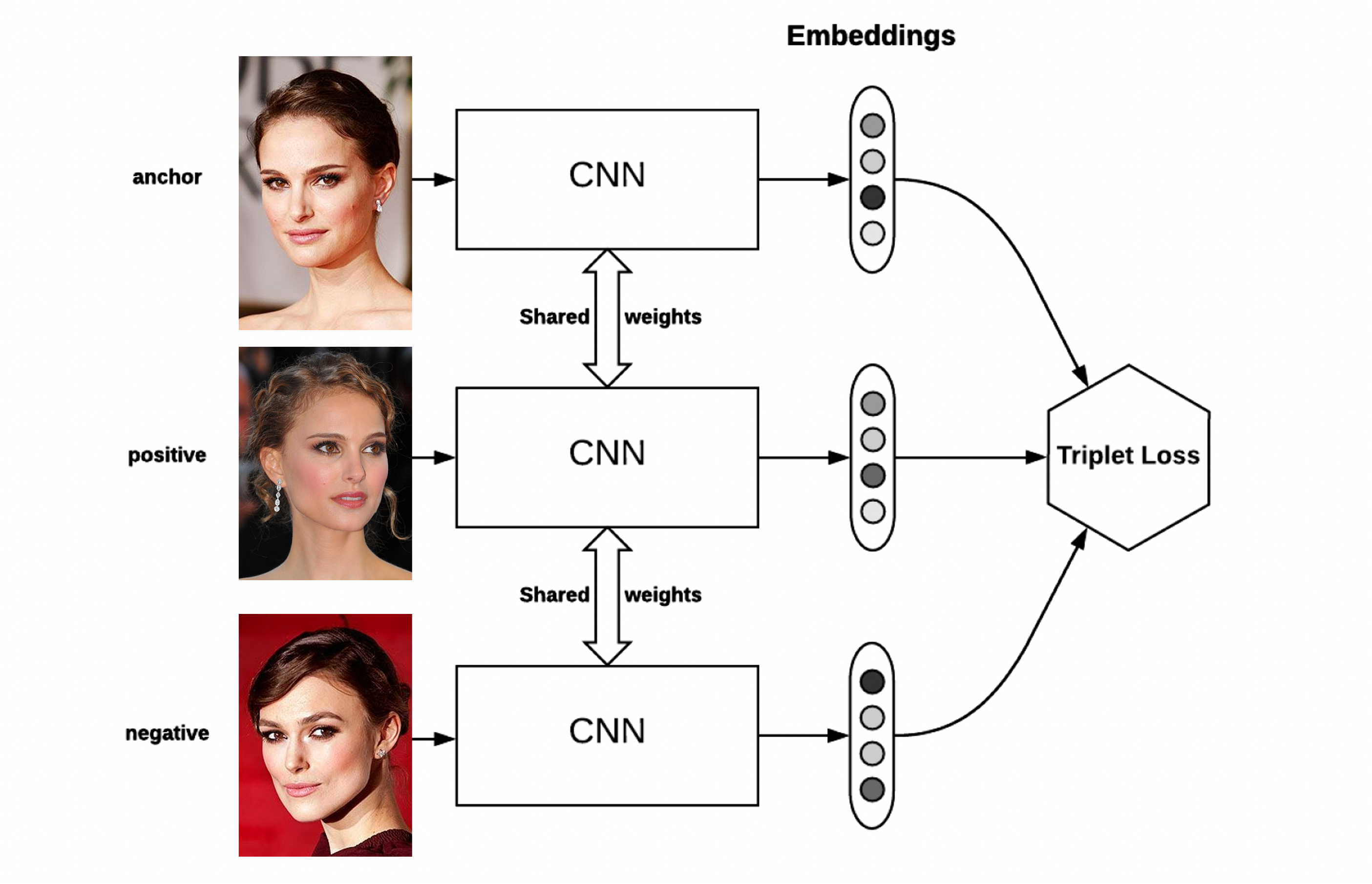}
    \caption{A possible example of hard triplets.}
    \label{fig:triplets}
\end{figure*}

\textcolor{black}{\subsection{Training Methodology}}
\textcolor{black}{The aim of the experiments was to build a strategy to better generalize the results beyond the training classes. For this reason, the goal of the training phase was to build a feature extractor rather than a classifier, i.e., to correctly position high-dimensional embeddings in a space in such a way that embeddings of the same animal are closer than embeddings of different ones. Furthermore, all training/testing partitions, as in the compared works, are animal-based, i.e., the same animal never appears in both the training and testing sets. 
For a network to learn to correctly position embeddings in a space, a type of contrastive loss is needed. The present study exploits the triplet loss instead of the cross-entropy loss, using a Siamese network (two identical networks with contrastive loss, or three identical networks with triplet loss) with either FaceNet or ViT. }
\textcolor{black}{For each element of the training set, a triplet is created. The embedding of the element itself is the anchor; then there are embeddings computed from a positive sample (same class as the anchor) and a negative sample (different class from the anchor). A network learns properly when it is fed hard examples, where the outcome is not clear. In other words, the more similar the negative sample to the anchor, possibly even more similar than the positive sample, the better. Figure \ref{fig:triplets} shows an example of hard triplets.\\}
\textcolor{black}{Finding the best triplets is a problem in itself. The implemented training procedure implements online triplet mining (during the processing of each batch of training data).
Since using only hard triplet mining could make the triplets too difficult for the network to learn, semi-hard triplet mining was used during the first epochs. In semi-hard mining, instead of looking at all the triplets, the network only focuses on those satisfying the ordering in the following equation (\ref{e:ordering}):
\begin{equation}
    \label{e:ordering}
    \|f(x_{i}^a) - f(x_{i}^p)\| < \|f(x_{i}^a) - f(x_{i}^n)\| < \alpha
\end{equation}
Together with the number of epochs, the value for the hyperparameter $\alpha$ was chosen empirically.\\
Hard triplet mining was used in the last phases: a hard triplet is a triplet in which the positive sample is farther from the anchor than the negative one, so that the outcome based on the distances would be wrong. The number of epochs to start hard mining is a further parameter that was experimentally investigated.}\\
\textcolor{black}{Regarding the possible cross-validation, the experiments on the different datasets followed the same strategies as the compared papers. When appropriate, this point will be detailed in reporting the evaluation results.}\\
\textcolor{black}{\textbf{Transfer learning from FaceNet.} In the presented approach, most of the network is frozen. Training only involves the layers starting from the last InceptionResNet block 
(Inception-resnet-C in Figure \ref{fig:FaceNet}).
The architecture used does not include a softmax layer because the network must not act as a classifier but rather as a feature extractor. The output consists of 512-dimensional embeddings that represent the input pictures.  }\\
\textcolor{black}{\textbf{Transfer learning from ViT.} To implement transfer learning for this work, we took Google's pre-trained weights on ImageNet-21k and updated them during training. Therefore, the training is from scratch, but with non-randomized weights. }\\

\section{Experimental Framework}
\label{s:evaluation}


The aim of the experiments is to assess whether the proposed approach can achieve SOTA performance, which is achieved by all existing proposals that exploit dedicated architectures and learning strategies. Of course, a fair comparison 
requires setting up the testing phase in exactly the same way as compared works, in particular using not only the same datasets, but also the same partitions, when indicated, and the same protocols, and to compute the same metrics of \cite{dogfacenet} and \cite{app11052074} for dogs, \cite{primates} for primates, \cite{bakhshayeshi2023intelligence}, \cite{hu2025angusrecnet} and \cite{hu2025two} for cattle.  
Therefore, except for the training of the chosen models, the performance evaluation replicates the setup described in the reference works. 
The proposed models are referred to as DogCNN (based on FaceNet) and DogViT when retrained on dogs, PrimCNN (based on FaceNet) and PrimViT on primates, and CattleCNN and CattleViT on cattle. \\
\subsection{Training Setup}
\textcolor{black}{Due to the differences among works and datasets, for the sake of the readers, it seems more appropriate to describe specific details of the related experimental setups in terms of data partition and cross-validation when reporting the results for the separate datasets. For backbone training, different parameter combinations were tested. In particular, the variations involved the number of epochs to pass from semi-hard to hard triplet mining (from 6 to 10); the number of epochs (from 30 to 50); patience early stopping, i.e., the number of epochs to wait before early stopping of training if no progress is achieved on the validation set (from 0 to 10); the parameter $\alpha$ in (\ref{e:ordering}) (from 0.1 to 0.3); and, finally, the number $P$ of identities and the number $K$ of images per identity in each batch, so that the batch size is $P \times K$ ($P = 16$ and $K = 2$ or $P = 8$ and $K = 4$). It is interesting that for both FaceNet and ViT consistently good results were achieved when the number of epochs before hard triplet mining was 6 (as was easy to expect due to the harder constraints), the best number of epochs was 50, the $patience$ parameter was 10, and $\alpha$ was 0.2; the number $K$ of images per individual per batch was consistently 4, while the number $P$ of individuals per batch was 8 for cattle and 24 or 16 for dogs, when using FaceNet or ViT respectively.}
When appropriate, we report the results of changing the $patience$ hyperparameter. In the case of the cattle, a preliminary hyperparameter search was performed using Keras’s Hyperband Tuner\footnote{https://keras.io/keras\_tuner/api/tuners/hyperband/}, and then training was carried out with those parameters and their variations.\\
\subsection{Evaluation Metrics}
\textcolor{black}{For the sake of readers not familiar with biometric systems, i.e., those systems designed to recognize single individuals based on selected physical or behavioral traits, it is worth introducing some basic definitions before reporting the evaluation results. First and foremost, biometric recognition is used in two applications: verification and identification. Reference templates of subjects to be recognized (\textit{enrolled}) are collected in the system \textit{gallery}}. \\
\textcolor{black}{\textbf{Biometric Verification. }In verification, when a probe is submitted to the system, there is either an explicit claim of identity by the subject to be recognized, or an implicit assumption (e.g., when authenticating on one's smartphone, the system assumes the owner's identity). In this case, after extracting the probe template (in the present cases, the embedding returned by the model) from the incoming sample, the comparison only involves gallery templates/embeddings belonging to the identity to be verified. The main performance metrics in this case are the False Acceptance Rate (FAR), i.e., the percentage of impostors (claiming an incorrect identity) that are wrongly accepted, and the False Rejection Rate (FRR), i.e., the percentage of genuine claims that are wrongly rejected. True Rejection Rate (TRR) and True Acceptance Rate (TAR) are the respective complements. All values are computed according to an acceptance threshold. The evaluation entails measuring the performance for different acceptance thresholds. Several metrics can be computed from these values (see \cite{jain2025introduction} for a detailed introduction). }\\
\textcolor{black}{We use the same verification metrics as the compared works. In particular, both works on dog recognition report verification results in terms of Accuracy (number of correct responses/total number of responses).  Verification results with primates consider the adopted 5-fold cross-validation, and are reported in terms of the mean ± s.d. of
the True Acceptance Rate (TAR) when FAR = 1\% across the 5 folds. Papers on cattle recognition do not report verification results. However, for the sake of completeness, we report those achieved by our models in terms of Accuracy, Precision, Recall, and the biometrics-specific  Equal Error Rate (EER), the error value achieved for that threshold where FRR = FAR.}\\
\textcolor{black}{\textbf{Biometric Identification. }In identification, there is no claim of identity. Therefore, it is up to the system to compare the probe template with all the templates stored in the gallery to determine the correct identity. An ordered list of candidates is produced. Regarding identification, it is due to further distinguish between closed.set (all probes are assumed to belong to an enrolled identity, so the only matter is to return the correct one in the first position, or rank, of the ordered candidate list), and open-set (there is a Reject option to classify the probe as belonging to an unknown/not enrolled identity). In closed-set identification, the most relevant metric is the Recognition Rate (RR), also referred to as Rank 1 recognition, or Cumulative Match Score at Rank 1 - CMS(1), i.e., the percentage of probes for which the correct identity was in the first position of the returned ordered list. Further, following rankings, typically Rank 5 and Rank 10, also called Cumulative Match Scores (CMS) at ranks 5 and 10, are also often reported in literature to assess how close to the list head is the correct identity. They respectively represent the percentages of probes for which the correct identity is returned in the list within position/rank 5 or within position/rank 10. The increasing CMS values are used to plot the Cumulative Match Characteristic (CMC) curve. The faster the curve grows, the better the system is. This motivates reporting percentages of correct identifications at rankings higher than the first one, but reasonably close to it. This is especially useful when a human expert further inspects a shortlist of candidates to manually identify the correct one. This happens in forensics \cite{dessimoz2008linkages}, but also in the case studies presented here to finally identify a missing pet (by the owner), a monitored animal (by a researcher), or an animal during feeding (by a farmer).\\}
\textcolor{black}{In identification open set, a threshold is involved to decide whether the probe belongs to an unknown subject, and then, if the threshold is met, the position of the correct identity in the ordered list of candidates is considered. Also for identification, a complete discussion of evaluation metrics can be found in \cite{jain2025introduction}.} \\
\textcolor{black}{In the present comparison, the metrics for identification closed set used for dogs are Recognition Rate (Rank 1) and Rank 5; for primates the results are reported in terms of mean ± s.d. of CMS at Rank 1 across the 5 folds; for cattles, the results are in terms of Rank 1 identification Accuracy (that we assume to be equal to CMS at Rank 1), Rank 5 and mAP (Mean Average Precision). The latter is the primary evaluation metric used to score and rank Re-ID system accuracy. It calculates the precision-recall curve, ensuring the model accurately retrieves the correct matches from a large image gallery. The only work reporting the results of identification open-set is the one on primate recognition. They are reported in terms of Detection and Identification Rate (DIR), i.e., the percentage of probes belonging to enrolled subjects that are correctly accepted and identified in the first list position, when FAR = 1\%.}\\

\textcolor{black}{It is important to underline that the metrics used in machine learning evaluation, especially Accuracy, are not completely appropriate for biometrics, where, especially for secure applications, the main goal is to avoid false acceptance. It is worth making a simple example. Let's assume a verification system with 100 incoming probes. Of these, 90 declare their true identity, while 10 declare a false one. Assuming that for a certain threshold the system accepts all claims, we would have 90\% Accuracy, which appears to be good performance. However, if we compute FAR and FRR for the same threshold, we find that the system achieved 100\% TAR, but also 100\% FAR (all impostors were accepted), which, especially in secure applications, is hardly acceptable. These considerations also motivate the study of optimal operational thresholds. The special focus on errors also motivates the FAR-oriented evaluation, in which an acceptable FAR is fixed, and systems are compared based on the TAR or FRR achieved at the corresponding threshold.  In the case of identification, a system that can return a reliable shortlist is often very useful in real-world applications. Accuracy only measures the correct results, but not the ability to quickly converge to the correct one. In any case, for a fair comparison, we will use the exact same metrics as the compared works.}\\

\subsection{Evaluation Results}
\subsubsection{Verification (1:1 identity comparison)}
%
\textbf{DogFace Dataset}. 
Verification experiments entailed pairing of samples that could produce either genuine pairs (images from the same subject) or impostor pairs (from different subjects). 
The tests followed the same protocol used in the compared works, using 100 draws each randomly selecting 2,500 genuine pairs and 2,500 impostor pairs from all possible pairs. 
In order to avoid bias possibly due to the particular choice of sample pairs, the
results represent the average performance from the 100 draws. 
As in the compared works, the performance is measured in terms of the best achieved accuracy. This value depends on the distance threshold chosen to consider a pair as genuine.
Table \ref{tabel:verification-table} 
shows 
that the compared SOTA model by Yoon at al. \cite{app11052074} reaches a mean accuracy of 88.8\%. The best model tested in this work, using ViT as a backbone, achieves a mean accuracy of 96.8\%, outperforming the old state of the art by 9\% and setting a new state of the art despite avoiding a dedicated architecture. The model based on FaceNet achieves only 88.2\%, which is in line with the state of the art.
%
\begin{table}[h]
\caption{\label{tabel:verification-table} Mean accuracy of each method on dogs dataset.}
\renewcommand{\arraystretch}{1.5} 
\scriptsize
\centering
\begin{tabular}{|c|c|c|} 
    \hline
    Learning Method  & Accuracy \\ 
    \hline\hline 
    DogFaceNet \cite{dogfacenet} & 76.9\% \\ 
    \hline
    Yoon et al. (SOTA) \cite{app11052074}  & 88.8\% \\ 
    \hline
    DogCNN  & 88.2\% \\ 
    \hline
    \textbf{DogViT}    & \textbf{96.8}\% \\ 
    \hline
 
\end{tabular}
\end{table}

\textbf{Primate Face Dataset}.
As already mentioned, the Primate Face Dataset actually includes three datasets. Since the testing framework is shared, it will be discussed jointly. 
Still following the same protocol of the compared work, the evaluation exploits 5-fold cross-validation. The verification performance is computed differently from DogFaceNet. 
For genuine comparisons, a probe image is compared with the same individual’s \textit{template} enrolled in the gallery, i.e., with all embeddings extracted from images of that individual, and the best result is taken as the outcome. Similarly, in the impostor comparisons, the probe is compared in turn with all the \textit{templates} from other individuals.
The probe/gallery split is repeated until every image has been used as a probe. The highest similarity score is selected when more gallery embeddings belong to the same individual. 
Verification results are the $mean \pm s.d.$ of the True Acceptance Rate (TAR) when FAR$=1\%$ across the 5 folds. Table \ref{t:ver-prim} shows the results: A stands for PrimCNN, B for PrimViT, and C for the compared PrimNet \cite{primates}. 

\textit{\underline{Lemur Face Dataset}}. The proposed methods do not outperform SOTA. PrimViT gets very close (with longer patience=10), yet with a considerably higher standard deviation. 
When this happens, depending on the chosen split, the performances may be better or worse, so it is not easy to determine which model is better. 
The PrimViT advantage lies in its quite close performance to SOTA, achieved through transfer learning rather than a dedicated architecture. 

\begin{table}[h!]
\caption{\label{t:ver-prim} Verification performance for the different primate groups expressed in terms of $mean \pm s.d.$ of TAR at FAR=1\%.}
\renewcommand{\arraystretch}{1.5} 
\scriptsize
\centering
\begin{tabular}{|c|c|c|c|}
    
    \hline
    \multicolumn{4}{|c|}{\textbf{Lemur Face}}\\
    \hline
    \multicolumn{1}{|c|} {Method} & \multicolumn{1}{|c|}{No re-training} & \multicolumn{1}{|c|} {Patience 1} & \multicolumn{1}{|c|}{Patience 10}\\ 
    \hline\hline\hline
    \multicolumn{1}{|c|}{A} &  \multicolumn{1}{|c|}{$63.30\% \pm 5.11\%$ }& \multicolumn{1}{|c|}{$67.35\% \pm 3.44\%$ }& \multicolumn{1}{|c|}{$67.22\% \pm 4.03\%$ } \\ 
    \hline
    \multicolumn{1}{|c|}{B} & \multicolumn{1}{|c|}{$72.60\% \pm 10.10\%$ }& \multicolumn{1}{|c|}{$73.7\% \pm 9.30\%$ }& \multicolumn{1}{|c|}{$81.70\% \pm 9.49\%$ } \\
    \hline
    \hline
    \multicolumn{1}{|c|}{C} & \multicolumn{3}{|c|}{\textbf{$83.11\% \pm 5.31\%$}}  \\ 
    \hline
    \multicolumn{4}{|c|}{\textbf{Golden Monkey Face}}\\
    \hline
    \multicolumn{1}{|c|}{Method} & \multicolumn{1}{|c|}{No re-training} & \multicolumn{1}{|c|}{Patience 1 }& \multicolumn{1}{|c|}{Patience 10}\\ 
    \hline\hline\hline
    \multicolumn{1}{|c|}{A} &  \multicolumn{1}{|c|}{$67.70\% \pm 8.50\%$ }& \multicolumn{1}{|c|}{$69.04\% \pm6.91\%$ }& \multicolumn{1}{|c|}{$70.54\% \pm 5.35\%$ } \\ 
    \hline
    \multicolumn{1}{|c|}{B} & \multicolumn{1}{|c|}{$52.60\% \pm 5.81\%$} & \multicolumn{1}{|c|}{$52.89\% \pm 6.90\%$ }& \multicolumn{1}{|c|}{$53.90\% \pm 7.50\%$ } \\
    \hline
    \hline
    \multicolumn{1}{|c|}{C} & \multicolumn{3}{|c|}{\textbf{$78.72\% \pm 5.80\%$}}  \\ 
    \hline
    \multicolumn{4}{|c|}{\textbf{Chimpanzees Face}}\\
    \hline
    \multicolumn{1}{|c|}{Method} & \multicolumn{1}{|c|}{No re-training} & \multicolumn{1}{|c|}{Patience 1} & \multicolumn{1}{|c|}{Patience 10}\\ 
    \hline\hline\hline
    \multicolumn{1}{|c|}{A} &  \multicolumn{1}{|c|}{$30.54\% \pm 8.87\%$ }& \multicolumn{1}{|c|}{$48.86\% \pm 6.35\%$ }& \multicolumn{1}{|c|}{$48.43\% \pm 7.31\%$ } \\ 
    \hline
    \multicolumn{1}{|c|}{B} & \multicolumn{1}{|c|}{$47.40\% \pm 7.01\%$} & \multicolumn{1}{|c|}{$60.81\% \pm 5.00\%$ }& \multicolumn{1}{|c|}{\textbf{$62.65\% \pm 5.51\%$}} \\
    \hline
    \hline
    \multicolumn{1}{|c|}{C} & \multicolumn{3}{|c|}{$59.87\% \pm 3.34\%$}  \\ 
    \hline
\end{tabular}
\end{table}

\textit{\underline{Golden Monkey Face Dataset}}. It is interesting to notice from Table \ref{t:ver-prim} that for this dataset, the best of the two proposed models is the simplest one, i.e., PrimCNN. Anyway, none of the proposed models could outperform SOTA.

\textit{\underline{Chimpanzee Face Dataset}}. For the Chimpanzee Dataset, the ViT-based model outperforms the state of the art, yet again with higher variance (Table \ref{t:ver-prim}).\\

\textbf{Cattle Dataset}.
Existing cattle face re-identification studies primarily focus on identification performance, where a query image is matched against a gallery of known individuals and results are reported in terms of recognition accuracy. For instance, the framework proposed by Bakhshayeshi et al. \cite{bakhshayeshi2023intelligence} reports identification accuracy in a query-to-gallery retrieval setup, whereas more recent approaches such as DCFuseViT and AngusRecNet similarly evaluate performance using re-identification or detection-oriented metrics. None of these works explicitly formulates the task as a biometric verification problem, i.e., a one-to-one (1:1) decision based on genuine and impostor comparisons with a learned decision threshold.
To address this limitation, in addition to the identification evaluation, this work introduces a dedicated biometric verification protocol based on embedding similarity distributions, threshold estimation through Equal Error Rate (EER) on a validation set, and subsequent evaluation on an independent test set. 
Under this protocol, the proposed CattleCNN model achieves 93.90\% accuracy, 98.71\% precision, and 98.71\% recall. For the CattleViT version, however, the performances report 99.42\% accuracy, 100\% precision, and 99.84\% recall. 
As for the EER values, they are 1.912\% for CattleCNN and 0.8196\% for CattleViT, respectively.

\subsubsection{Identification (1:N comparison)}
\textbf{Dog Face Dataset}.
The identification test with the dog dataset requires that the gallery contains one sample per identity in the test set (it is worth reminding that test identities do not appear in the training set), and the remaining images go in the probe.
The two compared works do not evaluate open set performance; therefore, the results here are consistently limited to closed set and reported in terms of Cumulative Match Scores (CMS) at Rank 1 and Rank 5. The data split for probe and gallery composition, and the subsequent tests, are repeated 1000 times.
Table \ref{tabel:RR-table} 
presents the results. 
%
DogViT outperforms the other proposed method. The most significant result is that it improves the current state of the art by Yoon et al. \cite{app11052074} of about 112\% at Rank 1, and of more than 40\% at Rank 5. 
In addition, the 96.9\% identification rate at Rank 5 enables the application of a shortlist-based search.



\begin{table}[h!]
\caption{\label{tabel:RR-table} Mean Cumulative Match Score at Rank 1 and 5 for each method on dogs.}
\renewcommand{\arraystretch}{1.5} 
\scriptsize
\centering
\begin{tabular}{|c|c|c| }
    \hline
    Learning Method & Rank 1 & Rank 5\\ 
    \hline\hline
    DogFaceNet \cite{dogfacenet} & 10.96\% & 32.58\%   \\ 
    \hline
     Yoon et al. (SOTA) \cite{app11052074} & 39.74\% & 68.80\%  \\ 
    \hline
    DogCNN &  38.19\% & 69.10\%  \\ 
    \hline
    \textbf{DogViT} & \textbf{84.34\%} & \textbf{96.97\%}  \\
    \hline
 
\end{tabular}
\end{table}

\textbf{Primate Face Dataset}.
Experiments include both closed- and open-set identification. In both cases, the evaluation is 5-fold, and there are 100 trials for the gallery/probe split to smooth out the randomness. A random sample
for every identity is chosen as a probe, and the rest are stored in the gallery. 
The results take into account the different performances of the 5 folds. Therefore, those for the identification closed-set are reported in terms of $mean \pm s.d.$ of CMS at Rank 1. The results of the identification open-set are reported in terms of $mean \pm s.d.$ of Detection and Identification Rate at Rank 1 when FAR$=1\%$. 
As above, in the following tables A stands for PrimCNN, B for PrimViT, and C for PrimNet \cite{primates}.

\textit{\underline{Lemur Face Dataset}}. In closed-set, the ViT-based proposed model outperforms SOTA, but with a (slightly) higher variance (Table \ref{tabel:primident}), while in open set the results are quite bad. 

\begin{sidewaystable}   
\caption{\label{tabel:primident}Performance of identification closed and open set for the different primate groups.}
\renewcommand{\arraystretch}{1.5} 
\scriptsize
\centering
\begin{tabular}{|c|c|c|c||c|c|c|}
    \hline
    \multicolumn{7}{|c|}{\textbf{Lemur Face}}\\
    \hline
    \multicolumn{1}{|c|} {} & \multicolumn{3}{c|}{\textbf{Closed set ($mean \pm s.d.$ of CMS at Rank 1)}} &\multicolumn{3}{|c|}{\textbf{Open set ($mean \pm s.d.$ of DIR at FAR = 1\%)}}\\
    \hline
    \multicolumn{1}{|c|}{Method }& \multicolumn{1}{c|}{No re-training} & \multicolumn{1}{c|}{Patience 1}  & \multicolumn{1}{|c|}{Patience 10}& \multicolumn{1}{c|}{No re-training} & \multicolumn{1}{c|}{Patience 1}  & \multicolumn{1}{|c|}{Patience 10}\\
    \hline\hline\hline

    A &  $79.40\% \pm 2.40\%$ & $79.47\% \pm 2.89\%$ & $76.56\% \pm 1.21\%$  &  $11.90\% \pm 2.10\%$ & $11.49\% \pm 2.17\%$ & $10.76\% \pm 2.34\%$ \\ 
    \hline
    B & $94.40\% \pm 1.30\%$ & $93.80\% \pm 1.90\%$ & \textbf{$94.00\% \pm 2.20\%$ } & $17.30\% \pm 2.90\%$ & $16.90\% \pm 3.00\%$ & $17.60\% \pm 1.50\%$ \\
    \hline
    \hline
    \multicolumn{1}{|c|}{C} & \multicolumn{3}{|c|}{$93.76\% \pm 0.90\%$} & \multicolumn{3}{|c|}{\textbf{$81.73\% \pm 2.36\%$}} \\ 
    \hline
    \multicolumn{7}{|c|}{\textbf{Golden Monkey Face}}\\
    \hline
\multicolumn{1}{|c|} {} & \multicolumn{3}{c|}{\textbf{Closed set ($mean \pm s.d.$ of CMS at Rank 1)}} &\multicolumn{3}{|c|}{\textbf{Open set ($mean \pm s.d.$ of DIR at FAR = 1\%)}}\\
    \hline
    \multicolumn{1}{|c|}{Method }& \multicolumn{1}{c|}{No re-training} & \multicolumn{1}{c|}{Patience 1}  & \multicolumn{1}{|c|}{Patience 10}& \multicolumn{1}{c|}{No re-training} & \multicolumn{1}{c|}{Patience 1}  & \multicolumn{1}{|c|}{Patience 10}\\
    \hline\hline\hline
    A &  $77.90\% \pm 3.30\%$ & $69.41\% \pm 8.56\%$ & $71.54\% \pm 6.01\%$ & $13.71\% \pm 5.22\%$ & $11.31\% \pm 6.05\%$ & $15.95\% \pm 7.22\%$ \\ 
    \hline
    B & $87.10\% \pm 5.06\%$ & $87.1\% \pm 7.02\%$ & \textbf{$92.80\% \pm 2.90\%$ } & $40.50\% \pm 11.20\%$ & $41.71\% \pm 11.59\%$ & $50.10\% \pm 11.89\%$\\
    \hline
    \hline
    \multicolumn{1}{|c|}{C} & \multicolumn{3}{|c|}{$90.36\% \pm 0.92\%$} & \multicolumn{3}{|c|}{\textbf{$66.11\% \pm 7.99\%$}}  \\ 
    \hline
    \multicolumn{7}{|c|}{\textbf{Chimpanzees Face}}\\
    \hline
    \multicolumn{1}{|c|} {} & \multicolumn{3}{c|}{\textbf{Closed set ($mean \pm s.d.$ of CMS at Rank 1)}} &\multicolumn{3}{|c|}{\textbf{Open set ($mean \pm s.d.$ of DIR at FAR = 1\%)}}\\
    \hline
    \multicolumn{1}{|c|}{Method }& \multicolumn{1}{c|}{No re-training} & \multicolumn{1}{c|}{Patience 1}  & \multicolumn{1}{|c|}{Patience 10}& \multicolumn{1}{c|}{No re-training} & \multicolumn{1}{c|}{Patience 1}  & \multicolumn{1}{|c|}{Patience 10}\\
    \hline\hline\hline
    A &  $53.44\% \pm 5.58\%$ & $52.95\% \pm 6.01\%$ & $48.04\% \pm 7.02\%$ & $1.25\% \pm 0.49\%$ & $7.05\% \pm 1.42\%$ & $7.26\% \pm 1.13\%$ \\ 
    \hline
    B & $64.40\% \pm 5.10\%$ & $73.95\% \pm 2.79\%$ & \textbf{$75.95\% \pm 3.57\%$} & $8.50\% \pm 2.10\%$ & $14.99\% \pm 3.96\%$ & $15.93\% \pm 3.35\%$ \\
    \hline
    \hline
    \multicolumn{1}{|c|}{C} & \multicolumn{3}{|c|}{$75.82\% \pm 1.25\%$} & \multicolumn{3}{|c|}{\textbf{$37.08\% \pm 11.22\%$} } \\ 
    \hline
\end{tabular}
\end{sidewaystable}
%
%
\textit{\underline{Golden Monkey Face Dataset}}. Table \ref{tabel:primident} further confirms the trend seen so far. 
Results are higher than SOTA for identification closed-set with ViT backbone, with a (slightly) higher variance. Different from lemurs, in the open-set, the results on golden monkeys are closer to the state of the art, but still considerably worse. It is also worth noticing that even the SOTA performances are low. 


\textit{\underline{Chimpanzee Face Dataset}}. Table \ref{tabel:primident} shows that, also for the Chimpanzees, the performance of PrimViT in closed-set tasks is slightly higher than SOTA, but with a higher variance. On the contrary, open-set performances of PrimViT are subpar; however, those in the compared work are also quite low.\\

\textbf{Cattle Dataset}
The one-to-many (1:N) evaluation performed in this work follows the re-identification commonly adopted in previous cattle face recognition studies, i.e., a closed set identification, where a query image is matched against a gallery of known individuals and the most similar identity is retrieved. This evaluation setup is consistent with the protocol introduced by Bakhshayeshi et al.\cite{bakhshayeshi2023intelligence}. Identification accuracy is based on query-to-gallery matching, as in subsequent works such as AngusRecNet and DCFuseViT that also focus on identity retrieval performance.\\
In the context of re-identification literature, the identification accuracy reported in previous cattle recognition studies corresponds to the Cumulative Matching Characteristic (CMC) Rank-1 metric, which measures whether the correct identity is retrieved at the top of the ranking. Since CMC Rank-1 is adopted as the primary metric in the compared works, the experiments here use it as well for fair comparison. \textcolor{black}{The comparison involves the best results reported in the corresponding studies.} While prior works mainly report only top-1 performance, the evaluation in this study is extended by including additional ranking-based metrics such as CMC Rank-5 and mean Average Precision (mAP), which provide a more comprehensive assessment of the quality of the embedding space and the overall ranking behavior. This enriched evaluation allows the analysis not only of whether the correct identity is retrieved first, but also of how consistently correct identities appear among the highest-ranked candidates. In both CattleCNN and CattleViT, the CMC Rank-5 value is 100\%, while for mAP the values are 98.71\% and 99.83\%, respectively.\\
Table \ref{tab:identification_cattle} summarizes the comparison between the proposed approach and the current state of the art.
\begin{table}[h!]
\begin{tabular}{|c|c|c|c|}
\hline
\multicolumn{1}{|c|}{Learning method} & \begin{tabular}[c]{@{}c@{}}Rank-1\\ Identification\\ Accuracy\end{tabular} & mAP & Rank-5  \\ \hline
Bakhshayeshi et a. \cite{bakhshayeshi2023intelligence}              & 95.13\%                                                                    & -    & -   \\ \hline

AngusRecNet \cite{hu2025angusrecnet}              & -                                                                          & 95.7\%  & -\\ \hline
DCFuseViT \cite{hu2025two}              & 96.5\%                                                                     & -     & -  \\ \hline
CattleCNN                             & 93.90\%                                                                    & 98.71\%  & 100\%   \\ \hline
\textbf{CattleViT}                             & \textbf{99.42\%}                                                                    & \textbf{99.83\%} & 100\%  \\ \hline
\end{tabular}
\caption{Identification performance for Cattely Dataset.}
\label{tab:identification_cattle}
\end{table}

\subsection{Preliminary Qualitative Analysis of Failures.}
Interestingly, image quality impacts the ability to leverage transfer learning. The datasets used exhibit decreasing quality, including lower resolution, potential blurring, and pose distortions. The best images are from the dog and cattle dataset, although they were acquired 'in the wild', whereas the primate dataset exhibits much lower quality. Indeed, the primate dataset features images acquired under much less favorable conditions, which certainly calls for more specific strategies. Interestingly, the transfer learning strategy's performance is in line with the quality of the datasets, with impressive results for dogs and cattle, both above SOTA, and inconsistent results for primates. This requires further investigation.\\

\begin{figure}[h!]
\centering
 \includegraphics[width=4cm]{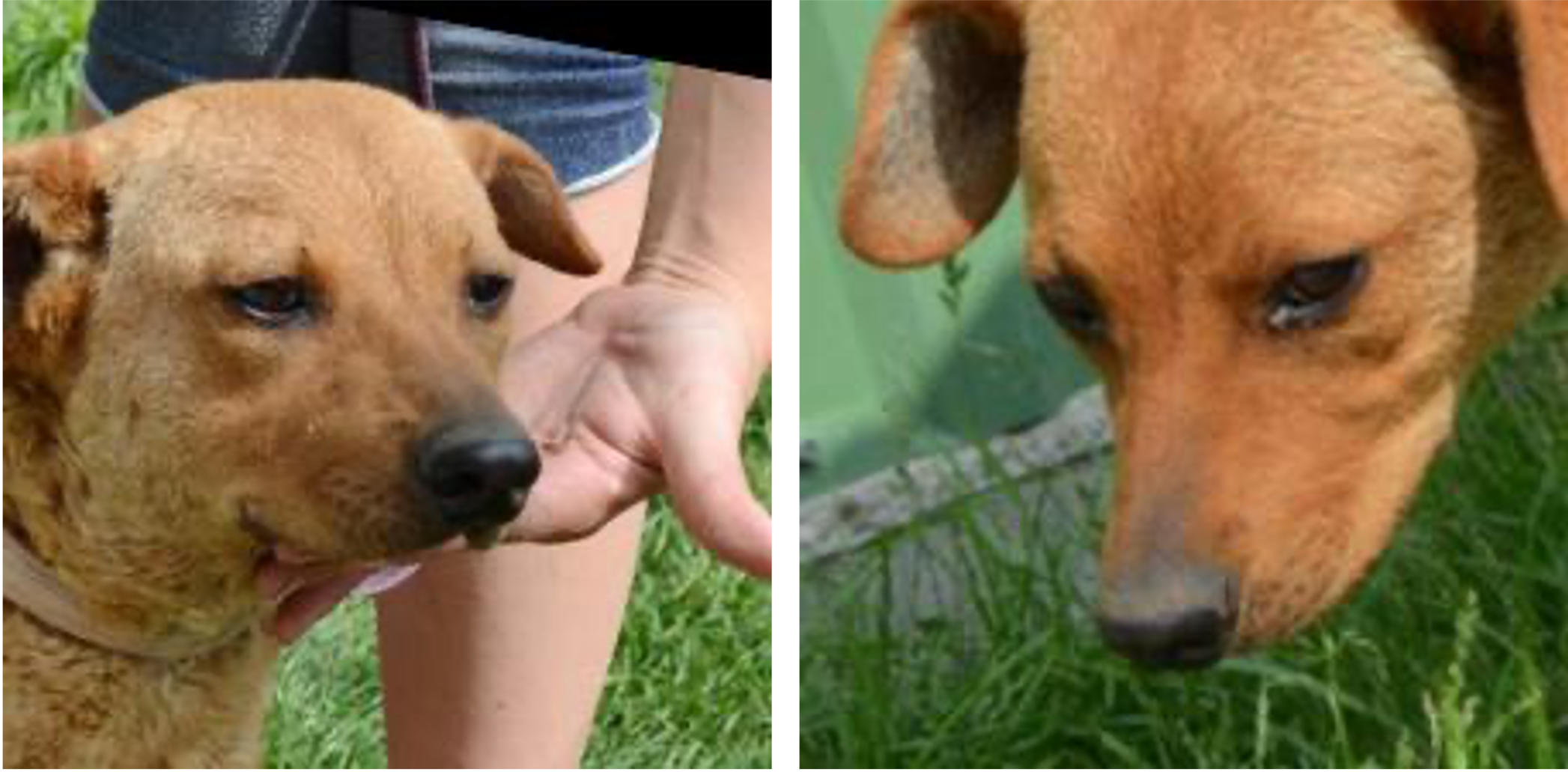}
   \caption{A false rejection for the dog dataset.}
    \label{fig:falserej}
\end{figure}

\begin{figure}[h!]
\centering
 \includegraphics[width=4cm]{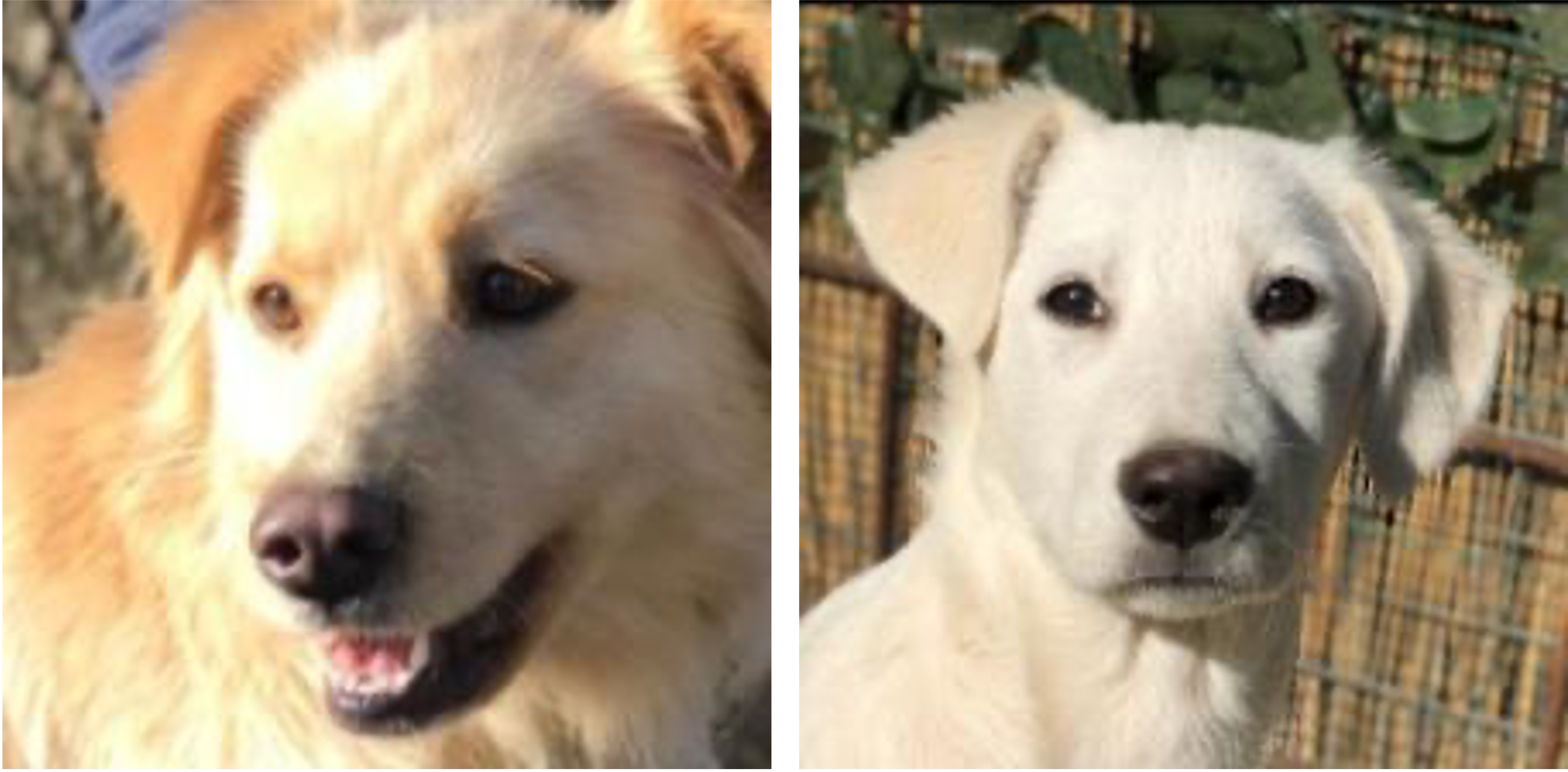}
   \caption{A false acceptance for the dog dataset.}
    \label{fig:falseacc}
\end{figure}

\begin{figure*}[h!]
\centering
 \includegraphics[width=11cm]{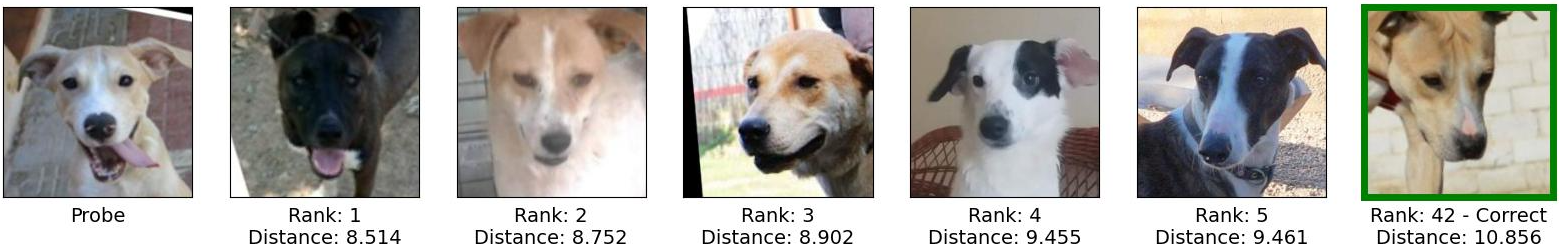}
   \caption{On the left: a probe image of a dog face; on the right: the top-5 retrievals plus the correct dog that is framed at rank 42.}
    \label{fig:wronglistdog}
\end{figure*}

\begin{figure*}[h!]
\centering
 \includegraphics[width=11cm]{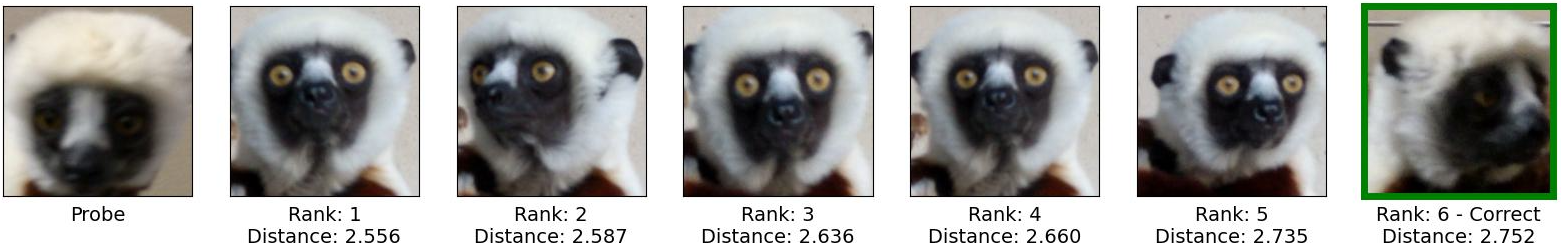}
   \caption{On the left: a probe image of a lemur face; on the right: the top-5 retrievals (the correct lemur is framed at rank 6.}
    \label{fig:wronglistlemur}
\end{figure*}

An analysis of misclassified dog pairs revealed some interesting clues. Figure \ref{fig:falserej} shows an example of a false rejection. It seems that pose is a crucial issue for animals as well, particularly the amount of pitch, given the generally prominent structure of the muzzle.
Figure \ref{fig:falseacc} shows an example of false accept. It seems that similar yet different colors get confused, causing an incorrect evaluation of overall similarity.
The previous observation is enforced by considering wrong identification rankings. In Figure \ref{fig:wronglistdog}, the correct animal appears at Rank 42. The one in the third position may appear similar. What deserves further consideration is the very first position occupied by a dog with a completely different color, as well as the ones in the fourth and fifth positions, which also appear much before the correct one. The same happens with primate images. Blurring is also a big issue, especially with images extracted from videos. Figure \ref{fig:wronglistlemur} shows a wrong ranking, where the images for all the first five ranks belong to the same lemur individual, different from the probe one.

\begin{figure*}[h!]
\centering
 \includegraphics[width=11cm]{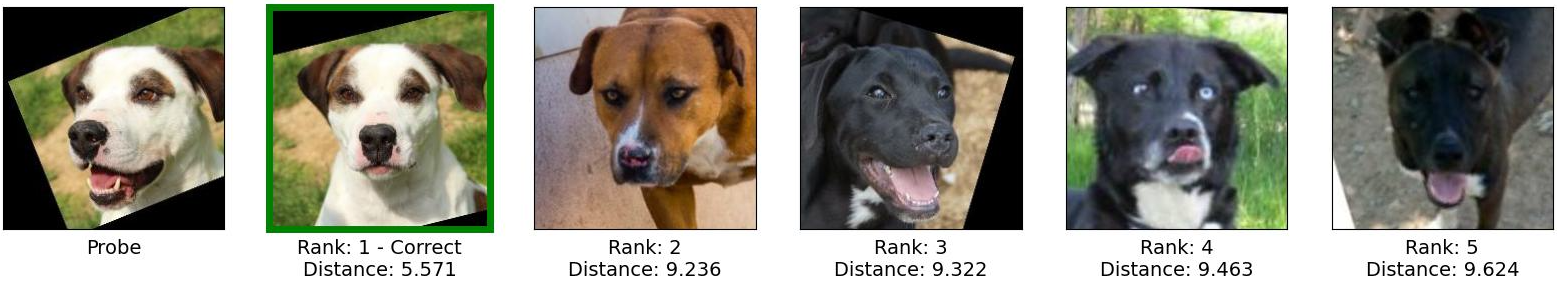}
   \caption{On the left: a probe image of a dog face; on the right: the top-5 retrievals (the correct dog is framed at rank 1).}
    \label{fig:oklistdog}
\end{figure*}

\begin{figure*}[h!]
\centering
 \includegraphics[width=11cm]{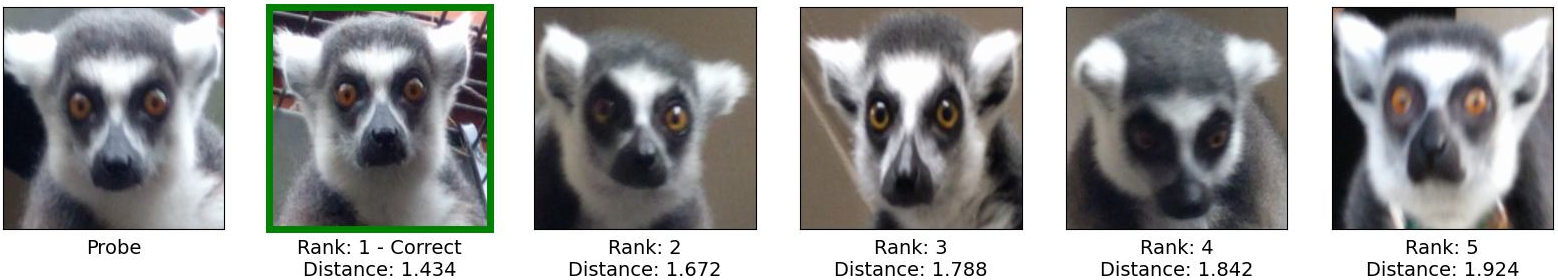}
   \caption{On the left: a probe image of a lemur face; on the right: the top-5 retrievals (the correct lemur is framed at rank 1).}
    \label{fig:oklistlemur}
\end{figure*}

The same issues affect correct rankings (Figures \ref{fig:oklistdog} and \ref{fig:oklistlemur}). In Figure \ref{fig:oklistdog}, the images after Rank 1 show dogs with a completely different color. Finally, it is interesting to notice the much lower distances among lemurs with respect to dog individuals (see distances in Figures \ref{fig:wronglistdog}, \ref{fig:wronglistlemur}, \ref{fig:oklistdog}, and \ref{fig:oklistlemur}). In Figure \ref{fig:oklistlemur}, Rank 1 image belongs to the correct lemur, but the following ones are very similar, and each of them belongs to a different lemur. This is particularly disruptive in identification, where the wrong individual at Rank 1 may actually be very close to the correct one. In open-set identification, this can increase the number of false accepts when an unregistered individual is mistakenly identified as a registered one, and false rejects when the correct individual is not returned at Rank 1.

\begin{figure}[h!]
\centering
 \includegraphics[width=4cm]{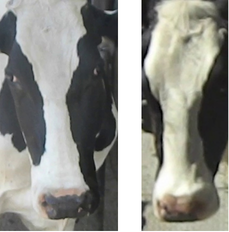}
   \caption{A false acceptance for the cattle dataset.}
    \label{fig:falseaccept_cows}
\end{figure}

An example of a false positive in the classification of cattle is also reported (Figure \ref{fig:falseaccept_cows}). As can be observed, the two animals share very similar morphological characteristics; therefore, it is plausible that the system identified them as the same and attributed the observed differences solely to variations in lighting conditions, which may also be expected to affect the appearance of colors. As noted in the case of dogs, lighting again proves to be a potentially decisive factor in the identification process. Also in that context, in fact, a false positive occurred under different lighting conditions, suggesting that the model may be particularly sensitive to such environmental variations.

\section{Conclusions and Future Work}
\label{s:conclusions}
This work explored a transfer-learning approach to animal face recognition, testing two backbones and three datasets with different image quality.
So far, most methods related to animal face recognition have used complex dedicated networks trained from scratch, e.g., for dogs. The present proposal investigates how to address the lack of data that negatively affects training by leveraging well-established networks designed for different tasks: human face recognition and object recognition. In addition, tests are made on three datasets with different image quality.
The approach seems very promising, but its real strength is in the possibility
to work with less data and with fewer weights to train, making learning lighter and faster. \\

%
The experiments compared the retraining of a section of an InceptionResNet used for human faces and of Vision Transformer (ViT) used for object recognition. The surprising outcome is that using ViT as a backbone, trained for object classification, achieves generally better results than FaceNet, expressly trained on (human) faces. This also happens despite the different image distributions.
In fact, the ViT-based approach achieves outstanding results over the old state of the art for dogs and cattle. However, it is not possible to say the same about endangered primates 
because of the higher standard deviation with respect to the compared work. This probably depends on the different nature of the datasets.
For instance, LemurFace contains 3000 images of 129 lemurs, while DogFace has more than 8300 images of 1393 dogs, with a very different number of samples per individual. ViT-based approach achieves outstanding results over the current SOTA on dogs, demonstrating that transfer learning from the transformer architecture can also learn to adapt to the data distributions at hand. A similar outcome can be observed in the cattle identification scenario, although the performance differences among the evaluated approaches are comparatively smaller. Even though it is not possible to say the same about all endangered primates, the results are, anyway, better than those achieved by the CNN-based transfer learning. In this context, the ViT-based method consistently achieves superior results, achieving near-perfect identification performance in most cases and demonstrating strong discriminative capability in the learned embedding space. The experiments also demonstrate that, when image quality decreases, as in the primate dataset, the transfer learning strategy is compromised and rarely achieves SOTA performance.\\

There are two points of possible criticism that deserve comment.\\
It is possible to observe that ViT is pre-trained on ImageNet, which contains more than 100 different dog breeds and also many monkey classes, including lemurs and chimpanzees. However, this may explain a good breed classification, while the experiments presented in this paper work at the single individual level. \\
A better evaluation protocol for dogs would involve selecting impostor pairs from the faces of different dogs of the same breed. On the one hand, this is not possible given the present labeling of the dataset\textcolor{black}{, which lacks the indication of the breed.  On the other hand, aiming at comparison, it was important to use the same experimental protocol with the same settings of the compared works. In these conditions, we still achieve better results than SOTA.}\\ 
%
%

In future work, it would be interesting to use longer training to analyze whether this improves the achieved results, especially on primates and in general on lower-quality images. It could also be interesting to further investigate the generalization capability of the proposed approach by evaluating its classification performance on additional datasets, particularly those related to cattle and referenced in the state of the art, as well as on datasets belonging to other animal species. Last but not least, it will also be worth analyzing in more detail the misclassifications for the two models. Some complementary behaviors may suggest a fusion strategy to improve results both with respect to individual scores and their possible fusion. 

\bibliography{DogViT}

\end{document}